\documentclass[conference]{IEEEtran}
\IEEEoverridecommandlockouts
\usepackage{booktabs}
\usepackage{multirow}
\usepackage{array}
\usepackage{cite}
\usepackage{amsmath,amssymb,amsfonts}
\usepackage{algorithmic}
\usepackage{algorithm}
\usepackage{graphicx}
\usepackage{textcomp}
\usepackage{xcolor}
\usepackage{booktabs}
\usepackage{multirow}
\usepackage{url}
\usepackage{hyperref}
\usepackage{tikz}
\usepackage{makecell}   
\usepackage{graphicx}
\usepackage{subcaption}
\usepackage{caption}

\usepackage{pgfplots}
\pgfplotsset{compat=1.18}
\usetikzlibrary{arrows.meta, positioning, fit, backgrounds, shapes.geometric}

\def\BibTeX{{\rm B\kern-.05em{\sc i\kern-.025em b}\kern-.08em
    T\kern-.1667em\lower.7ex\hbox{E}\kern-.125emX}}

\begin{document}

\title{Real-Time Human Trajectory Prediction from RGB-D for Mobile Robots Using Bi-Mamba

\thanks{This research was supported by several NIT projects: in part by the Federal Ministry of Research, Technology, and Space of Germany (BMFTR) through project Edison (Grant No. 13N17576); in part by the European Regional Development Fund (ERDF) through ENABLING (Grant No. ZS/2023/12/182056); and by the ERDF as part of the project ORAKEL (Grant No. ZS/2023/12/182322), funded by the European Union and the state of Saxony-Anhalt. }
}

\author{\IEEEauthorblockN{1\textsuperscript{st}
Basheer Al-Tawil}
\IEEEauthorblockA{
\textit{Neuro Information Technology}\\
\textit{Otto von Guericke University }\\
39106 Magdeburg, Germany\\
basheer.al-tawil@ovgu.de}
\and
\IEEEauthorblockN{2\textsuperscript{nd} Magnus Jung}
\IEEEauthorblockA{
\textit{Neuro Information Technology}\\
 \textit{Otto von Guericke University }\\
 39106 Magdeburg, Germany\\
magnus.jung@ovgu.de}
\and
\IEEEauthorblockN{3\textsuperscript{rd} Ayoub Al-Hamadi}
\IEEEauthorblockA{
\textit{Neuro Information Technology}\\
\textit{Otto von Guericke University }\\
39106 Magdeburg, Germany
\\ ayoub.al-hamadi@ovgu.de}
}

\maketitle              
\begin{abstract}
Predicting the short-term future trajectory of a person from RGB-D cameras is a key capability for mobile robots navigating in dynamic and human populated environment. The key challenge to deploying learned predictors on real robots is the train-deploy pipeline gap. Training data are often prepared with an idealized, smoothed ground truth, whereas at inference, the model is fed noisy estimates from pose detectors and depth sensors. The model is therefore evaluated on an input distribution it never encountered during training. Most existing models rely on recurrent neural networks (RNNs) or transformers, which either struggle to capture short-term temporal dependencies or suffer from computational cost. We present TrajMamba, a lightweight, bidirectional state space model for short-term human trajectory prediction using an RGB-D camera. The model uses a single shared processing path for both training and deployment, aligning human pose extraction, depth back projection, and smoothing processes. It encodes a short window of persons' hip center velocities with stacked bidirectional Mamba blocks and decodes future velocities with a GRU head. On the ETH/UCY benchmark, TrajMamba reaches an average ADE of 0.29 \,m FDE of 0.51\,m. With 262K parameters, it runs in real time on modest hardware and is deployed on a TIAGo robot with an onboard NVIDIA Jetson Orin.

\end{abstract}

\section{Introduction}
\label{sec:intro}

\begin{figure*}[h!]
    \centerline{\includegraphics[width=\textwidth,clip,trim=0 22cm 0cm 0cm]{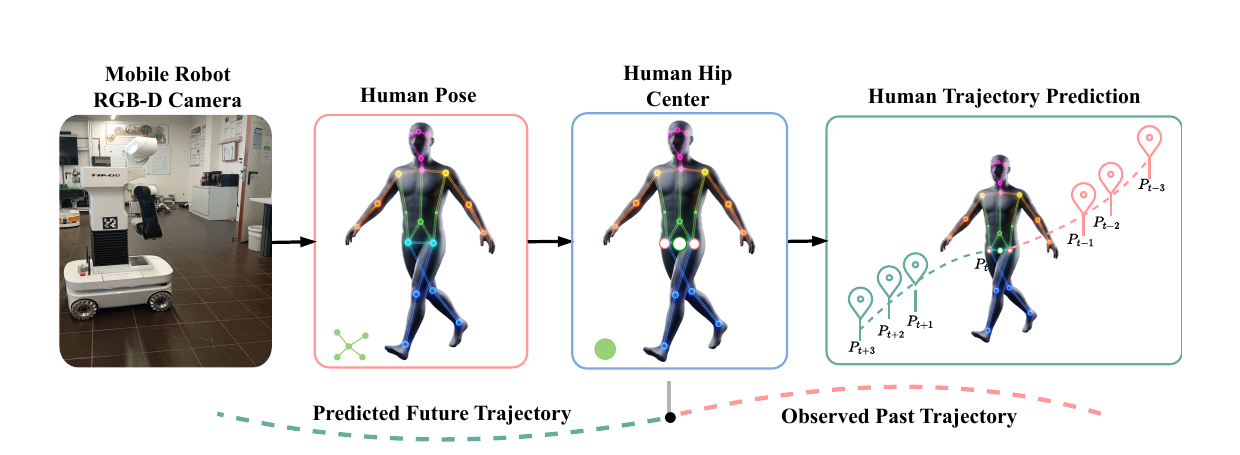}}
    \caption{Problem setting. A TIAGo PAL robot~\cite{pages2016tiago} mounted RGB-D camera observes a person. MediaPipe extracts the hip center from each frame, which is then back-projected to 3D using depth information. The goal is to estimate the person’s future trajectory from the observed hip positions.}
    \label{fig:problem_setting}
\end{figure*}
Trajectory estimation of a person over short, rapid time horizons is essential for collision-free robot navigation in dynamic environments~\cite{alahi2016social, al2025mobile}. For people and robots to interact safely in shared spaces, a robot's predicted trajectories for moving people must be consistent with their social behavior. Over the past decade, many data-driven systems have been presented for this task, ranging from early recurrent neural networks (RNNs)~\cite{alahi2016social}, to attention-based networks~\cite{huang2019stgat, al2025multi} and transformers~\cite{giuliari2021transformer,yuan2021agentformer}, and more recently to diffusion-based generative models~\cite{gu2022stochastic}. Most of these methods are evaluated on trajectory-based benchmarks derived from overhead video observations of pedestrian motion, such as ETH/UCY, or that assume clean, pre-computed skeletons~\cite{usman2022skeleton}. This setting does not match how a mobile robot actually sees the world. On a robot, the input comes from a forward-facing
RGB-D camera, and the person's position must be estimated on the fly from noisy pose and depth data~\cite{huang2025trajectory}. Figure~\ref{fig:problem_setting} illustrates the target problem addressed in this work.

This mismatch creates a practical but often overlooked problem. During training, the model usually learns from the clean, smoothed ground truth provided with a dataset.
At deployment, it is instead fed live estimates from a pose detector and a depth sensor, which carry jitter, noise, and occasional gaps. The model is therefore asked to work on a kind of input it never saw while learning. We call this the train-deploy pipeline gap. Our solution is to remove this gap by design. We use the same processing pipeline, MediaPipe pose extraction, depth back-projection, and Kalman/EMA filtering in both training and inference. On top of this, we build
TrajMamba, a small and fast bidirectional state space model that predicts a person's short-term trajectory and runs in real time on a robot. The complete system is released as an open-source ROS2 package, together with the training pipeline and
trained checkpoints. \footnote{\url{https://github.com/basheeraltawil/NIT-TrajMamba}}

Our main contributions are:
\begin{itemize} 
\item TrajMamba, a lightweight bidirectional Mamba-GRU model for real-time trajectory prediction ($\approx$262K parameters).
\item A pipeline-consistent strategy that shares the same pose, depth, and filtering code across training and inference, with normalization stored in the checkpoint, reducing the train-deploy gap.
\item Evaluation on ETH/UCY and real-time deployment on a TIAGo PAL robot~\cite{pages2016tiago}, NVIDIA Jetson Orin.

\end{itemize}

\section{Related Work}
\label{sec:related}

\subsection{Pedestrian Trajectory Prediction}

The Social LSTM model~\cite{alahi2016social} established a significant breakthrough in the vertical binary direction, and researchers have made new innovations in the development of this model through read-head encoders such as STGAT~\cite{huang2019stgat}, convolutional models~\cite{nikhil2018convolutional}, graph-based predictors
such as Social-STGCNN~\cite{mohamed2020social}, SR-LSTM~\cite{zhang2019sr} and advanced architectures (transformers) designed for scalability (such as Trajectron++~\cite{salzmann2020trajectron++} and AgentFormer~\cite{yuan2021agentformer}), as well as stochastic generative developments~\cite {gu2022stochastic}. Although these methods have emerged and demonstrated their versatility on vertical datasets, they operate exclusively in the 2D top-down plane and cannot be directly applied to the RGB-D camera-based environment in which they are embedded.
Our work builds upon this achievement: we focus on predictions of a person’s motion through the camera, prioritizing real-time performance and real-time efficiency, rather than sampling and ranking many stochastic futures.

\subsection{3D Motion Forecasting}

Research on predicting the 3D motion of an entire body, such as PoseForecaster~\cite{cao2020long}, often requires clean, pre-computed skeleton inputs, rather than dealing with raw sensor streams. Meanwhile, depth-based action recognition systems~\cite{wang2018videos} process full-depth sequences for classification rather than spatial prediction. 
On the other hand, Hasan et al.~\cite{hasan2018mx} predict 3D trajectories from the depth sensor, but they rely on background subtraction as a front-end for detection, a technique that is fragile and easily fails when the camera moves. 
In this work, we build our system on an advanced pose estimator, MediaPipe PoseLandmarker~\cite{bazarevsky2020blazepose}, and integrate it as a common component in both the training and inference stages.

Recent research in the field of 6-DOF object trajectory generation from a first-person perspective includes EgoFlow~\cite{saroha2026egoflow}, a flow matching technique integrated with bidirectional Mamba and a perceiver-based encoder to produce physically consistent object motion from multisensory observations. EgoFlow is designed to handle objects via wearable cameras and focuses on generation rather than predicting human motion. Its architectural vision, specifically the effectiveness of bi-directional Mamba in modeling temporal sequences, strongly supports our design choice to predict the trajectories of people.

\subsection{State Space Models for Sequence Modeling}
\label{sec:related_ssm}

Structured state space models have demonstrated that linear recurrences with carefully designed transition matrices can achieve competitive performance in modeling long sequences. The Mamba model~\cite {gu2023mamba} has extended this concept by introducing selective state spaces. It makes both state and input transition matrices content-dependent, supporting the model with data-adaptive memory without incurring the computational cost as in transformers. 
Mamba-based architectures have been successfully applied to trajectory prediction: Trajectory Mamba~\cite{liu2026trajmamba} achieved competitive results on pedestrian trajectory estimation. It strengthened research in multi-agent settings and demonstrates the model’s scalability and adaptability. Bidirectional Mamba processing, which fuses a forward and a backward pass through a learnable gate, has also been shown to benefit action and motion modeling in visual tasks~\cite {saroha2026egoflow,zhang2024motion}.
Our model inherits this bidirectional design and adapts it to the constraints of RGB-D robot perception by reducing the parameter count and sharing the entire sensor processing steps between training and deployment.

\section{Method}
\label{sec:method}

\subsection{Problem Formulation}
\label{sec:problem}

\begin{figure*}[h!]
    \centerline{\includegraphics[width=\textwidth,clip,trim=0 0cm 0 0cm]{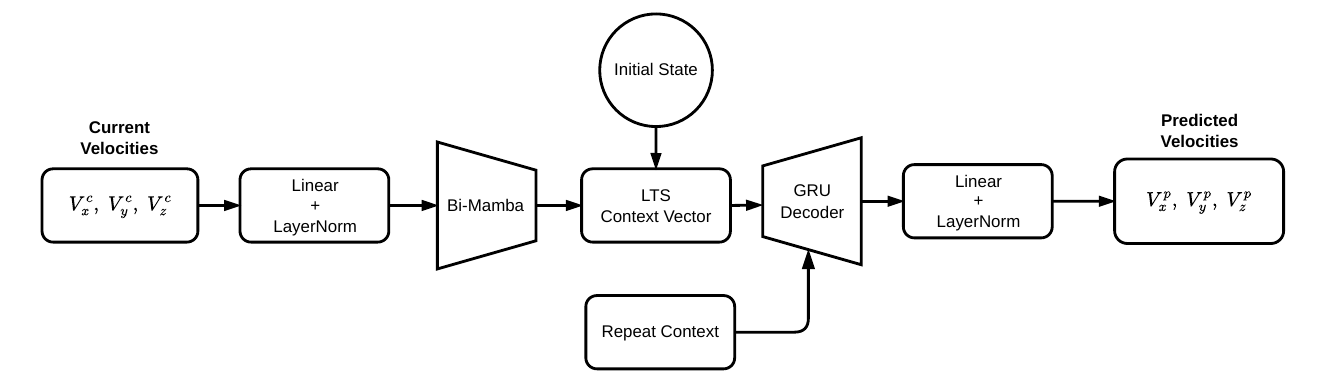}}
    \caption{ Framework structure. Normalized observed velocities pass through a
linear layer and several bidirectional Mamba blocks (forward + backward, merged
by a learned gate). The last time step (LTS) of the encoder initializes a GRU
decoder, whose outputs go through a small MLP to predict future normalized
velocities.}
    \label{fig:system arch}
\end{figure*}
Suppose that $\mathbf{p}_t \in \mathbb{R}^3$ represents the 3D position of the center of the person's hip in the camera coordinates at time step $t$.
Given the observation window 
$\mathcal{O} = \{\mathbf{p}_1, \ldots, \mathbf{p}_{T_\text{obs}}\}$. The goal is to predict the future trajectory 
$\mathcal{F} = \{\mathbf{p}_{T_\text{obs}+1}, \ldots, \mathbf{p}_{T_\text{obs}+T_\text{pred}}\}$ of a person based on RGB-D input frames.
Following common practice~\cite{alahi2016social}, we operate in \emph{velocity space}: where the model input is the first-order finite differences $\mathbf{v}_t = \mathbf{p}_{t+1} - \mathbf{p}_t$, and predicted positions are retrieved through cumulative summation. Velocities are standardized per spatial axis using the training-set mean $\boldsymbol{\mu}_v$ and standard deviation $\boldsymbol{\sigma}_v$. The complete architecture is shown in Figure \ref{fig:system arch}.

\subsection{Pipeline-Consistent and Pre-processing}
\label{sec:extraction}

\begin{figure}[h!]
    \centerline{\includegraphics[width=0.5\textwidth,clip,trim=0 0cm 0 0cm]{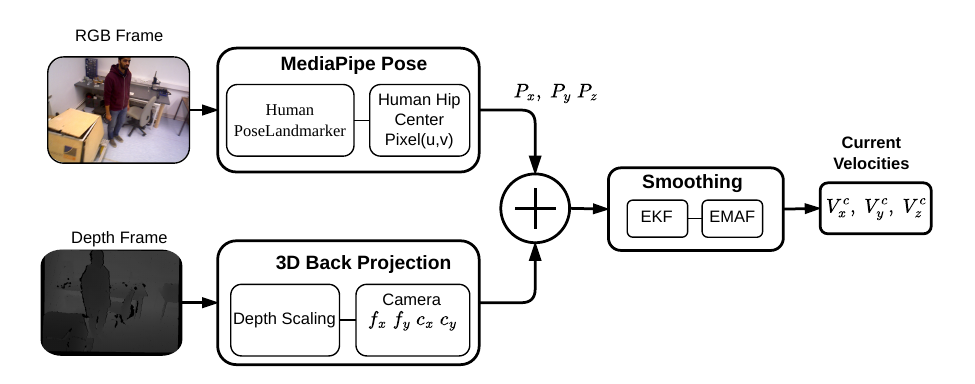}}
    \caption{Preprocessing: From raw RGB-D frames to normalized velocities. Frames are paired by timestamp, MediaPipe locates the hip, and depth gives 3D position. The signal is smoothed (Kalman + EMA), downsampled, and split into windows. Velocities are computed via finite differences and normalized.}
    \label{fig:Preprocessing}
\end{figure}

The core principle of our approach is that every processing step applied to sensor data during training must be reproduced exactly at inference time, as shown in Figure~\ref{fig:Preprocessing}. We therefore implement a single end-to-end module that is called in both phases. During the preprocessing of training and testing data and while the robot is operating in the real world. It handles all operations, including synchronization of row RGB-D frames, geometric body pose extraction, and conversion of image pixels to 3D spatial coordinates (back-projection). It applies smoothing filters (Kalman and EMA) to clean the information token. In addition, the unit divides the data into overlapping windows and mathematically normalizes it to suit the neural network. This structural design ensures that the model receives data processed with the same logic in both the training, testing, and real-world deployment processes.

\subsubsection{RGB-D Frame Association}
The features of the RGB-D frames, along with their timestamps, are extracted. Then, each color image is associated with its corresponding depth image. The association algorithm searches for the closest timestamp between them, provided that the time difference does not exceed 20 ms. This threshold mirrors the streaming behavior of the live RealSense camera, keeping synchronization consistent between offline data and on-robot operation.

\subsubsection{Pose Detection}

Each RGB frame is passed to MediaPipe PoseLandmarker in video mode, which activates the detector's internal timestamp-driven temporal smoothing without introducing additional model hyperparameters. From the normalized
landmark coordinates of the left hip ($\mathbf{l}_{23}$) and right hip ($\mathbf{l}_{24}$), each in $[0,1]$, we compute the hip center pixel $(u,v)$ by averaging the two and scaling by the image width $W$ and height $H$:
\begin{equation}
u = \frac{l_{23}^{x} + l_{24}^{x}}{2}\, W, \qquad
v = \frac{l_{23}^{y} + l_{24}^{y}}{2}\, H,
\label{eq:hipcenter}
\end{equation}
where $l^{x}$ and $l^{y}$ denote the normalized horizontal and vertical
coordinates of a landmark, respectively.

\subsubsection{Depth Sampling and Back-Projection}
To estimate the true distance $z$ of a person from the camera, the system does not rely on a single pixel reading. Instead, it extracts a $7\times7$ pixel patch centered around the bin coordinates $(u, v)$ from the 16-bit depth image. It discards zero values, which represent invalid or missing depth readings. Then, it calculates the median of the remaining values. These values are converted to meters by dividing it by the standard TUM scale factor (pixel value / 5000)~\cite{sturm2012benchmark}.
Once the depth $z$ is obtained in meters, the system applies the standard pinhole camera model to perform back-projection, converting the 2D pixel into 3D spatial coordinates in camera space, according to the following equation:
\begin{equation}
x = \frac{(u - c_x)\,z}{f_x}, \quad
y = \frac{(v - c_y)\,z}{f_y}, \quad
\mathbf{p} = [x,\; y,\; z]^\top.
\label{eq:backproject}
\end{equation}
To ensure the accuracy of this projection, the RealSense camera intrinsics are provided as ($f_x{=}384.4$, $f_y {=}384.4$, $c_x{=}321$, $c_y{=}231$).

\subsubsection{Smoothing}
Since the extracted 3D coordinates are noisy and have random vibrations, we pass them through a cascaded filtering structure consisting of two stages. First, a one-dimensional Kalman filter is applied independently to each motion axis ($x$, $y$, and $z$) based on a constant-velocity model, with process noise parameters $q{=10^{-3}}$ and measurement noise parameters $r{=10^{-2}}$ to isolate sensor noise. 
Second, the outputs are combined and passed directly through an exponential moving average filter (EMA Filter) with a coefficient $\alpha{=0.35}$ to dampen any remaining high-frequency oscillations.

\subsection{Sliding-Window Dataset Construction}
\label{sec:dataset_construct}
Extracted trajectories are downsampled by a factor of $3$ (keeping
every third frame),
reducing the rate from $\sim$30\, Hz to $\sim$10\, Hz to limit redundancy between
near-identical frames and ease the on-board compute load. The downsampled
tracks are split into overlapping windows of length
$T_\text{obs} + T_\text{pred} = 20 + 30 = 50$ frames, i.e.\ a $2$\,s observation horizon and
a $3$\,s prediction horizon. Each window is translated so its first frame sits at the origin, forcing the model to learn relative motion rather than absolute scene coordinates.

\subsection{Model Architecture}
\label{sec:arch}

Figure ~\ref{fig:system arch} illustrates the architectural structure of the model. It is designed to be compact and lightweight to accommodate the computational constraints of on-board robotics. It consists of a bidirectional Mamba encoder and decoder based on a recurrent neural network (GRU).

\subsubsection{Input Embedding}
The 3D velocity vector $\mathbf{v}_t \in \mathbb{R}^3$ is received and passed through a linear projection layer to project it into the $d_\text{model}$-dimensional semantic space, followed by a layer normalization step to normalize the tokens and ensure the stability of gradients during training.

\subsubsection{Bidirectional Mamba Encoder}

\begin{figure}[h]
    \centerline{\includegraphics[width=0.5\textwidth,clip,trim=0cm 0cm 0cm 0cm]{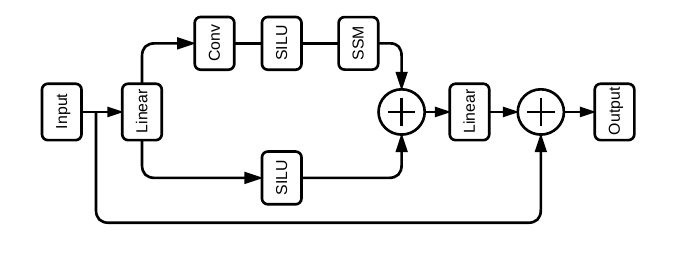}}
    \caption{Inside a Mamba block. The input splits into two branches: one goes through a depthwise convolution and the selective SSM , the other acts as a gate. The SSM uses the input dependent $\Delta$ to discretize a fixed HiPPO matrix $\mathbf{A}$, then runs a recurrent scan. The gated product is then projected and added to the input residual.}
    \label{fig:mamba block}
\end{figure}

The encoder stacks $N$ bidirectional Mamba blocks; see Figure~\ref{fig:mamba block}. To capture both past and future temporal context, every block processes the sequence in two
directions: a forward pass over the original order and a backward pass over the
time-reversed sequence. The two passes are fused by a learned per-channel gate:

\begin{equation}
\mathbf{h}_t = g_t \odot \mathbf{h}_t^{\text{fwd}}
+ (1 - g_t) \odot \mathbf{h}_t^{\text{bwd}},
\label{eq:gate}
\end{equation}
with $g_t = \sigma\!\big(\mathbf{W}[\mathbf{h}_t^{\text{fwd}};
\mathbf{h}_t^{\text{bwd}}]\big)$, followed by layer normalization. Each block
implements the selective space S6 mechanism of Mamba~\cite{gu2023mamba}, where the parameters $(\Delta_t, \mathbf{B}_t,
\mathbf{C}_t)$ are computed from the input by a shared projection, $\Delta_t$ is
passed through softplus to keep it positive, and the recurrence is discretized
via zero-order hold.

\begin{align}
\bar{\mathbf{A}}_t &= \exp(\Delta_t \cdot \mathbf{A}), \label{eq:dA}\\
\bar{\mathbf{B}}_t &= \Delta_t \cdot \mathbf{B}_t, \label{eq:dB}\\
\mathbf{h}_t &= \bar{\mathbf{A}}_t \odot \mathbf{h}_{t-1}
               + \bar{\mathbf{B}}_t \cdot \mathbf{x}_t, \label{eq:scan}\\
\mathbf{y}_t &= \mathbf{C}_t \cdot \mathbf{h}_t
               + \mathbf{D} \cdot \mathbf{x}_t. \label{eq:out}
\end{align}
The diagonal state matrix $\mathbf{A}$ is initialized from the HiPPO spectrum
(log-parameterized for stability) and refined during training.

\subsubsection{GRU Decoder}
The final encoder output $\mathbf{c}=\mathbf{h}^{\text{enc}}_{T_\text{obs} -1}$ is used
as a global context vector. It initializes the GRU hidden state through
$\mathbf{s}_0=\tanh(\mathbf{W}_c\mathbf{c})$ and is repeated as the input at
each of the $T_\text{pred}-1$ decoding steps, so every prediction is conditioned on
the full observation. A two-layer MLP with dropout maps each GRU output to a predicted future
velocity $\hat{\mathbf{v}}_t$.

\begin{figure*}[h!]
    \centering
    \setlength{\tabcolsep}{2pt}      
    \renewcommand{\arraystretch}{0.6}
    \begin{tabular}{ccccc}
        ETH & UNIV & Zara1 & Zara2 & Hotel \\
        \includegraphics[width=0.18\textwidth]   {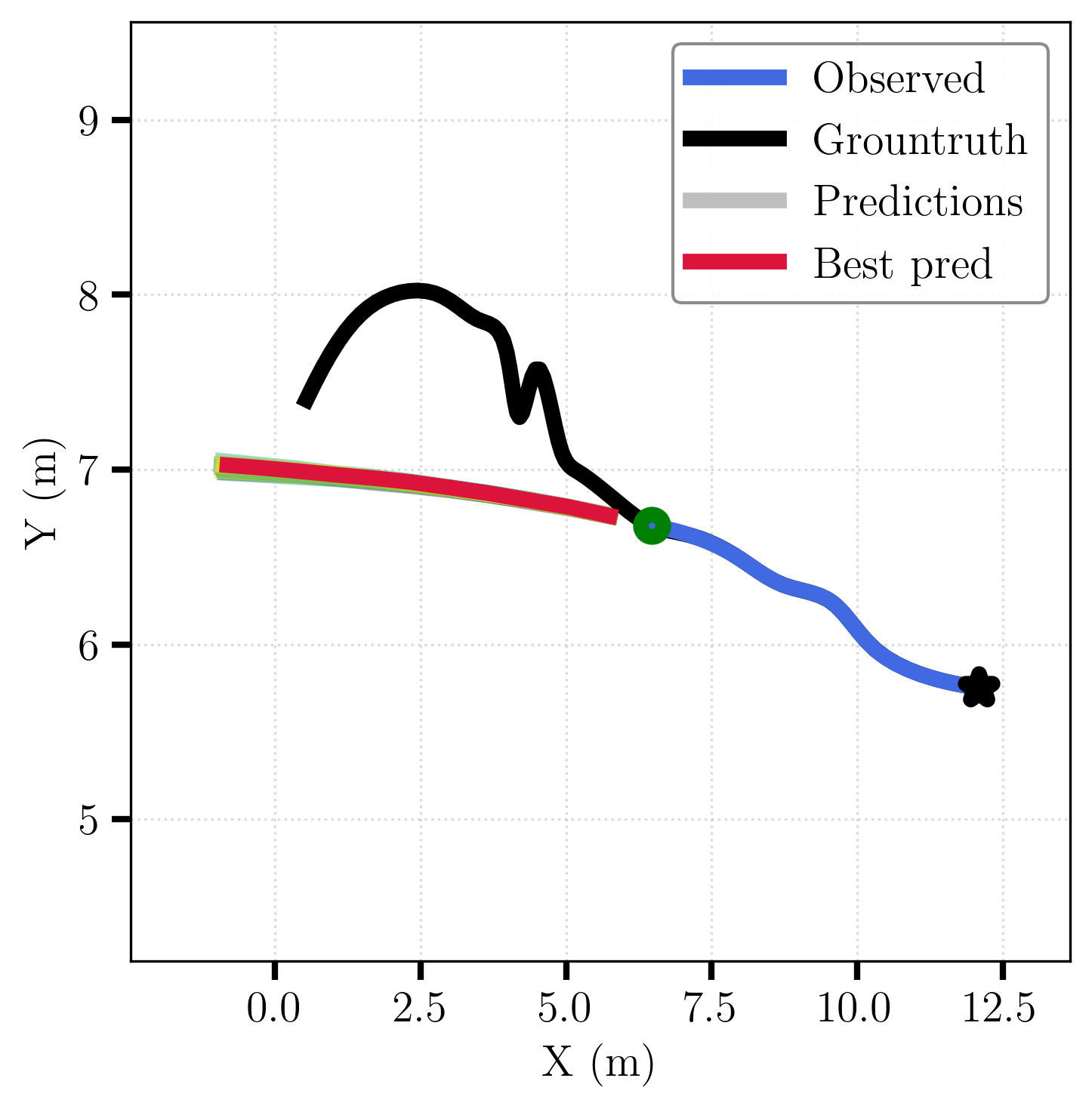}    &
        \includegraphics[width=0.19\textwidth]{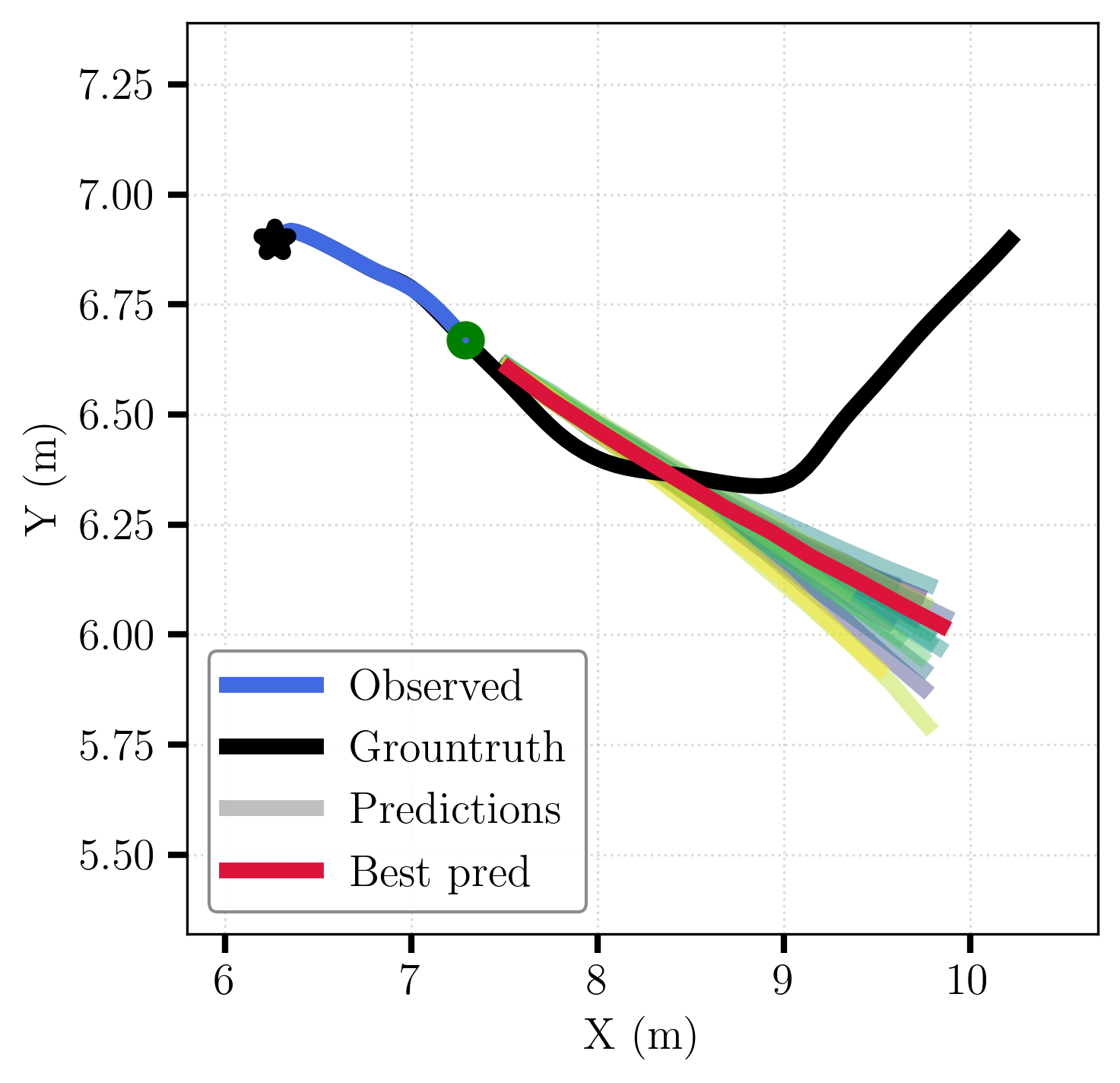} &
        \includegraphics[width=0.19\textwidth]{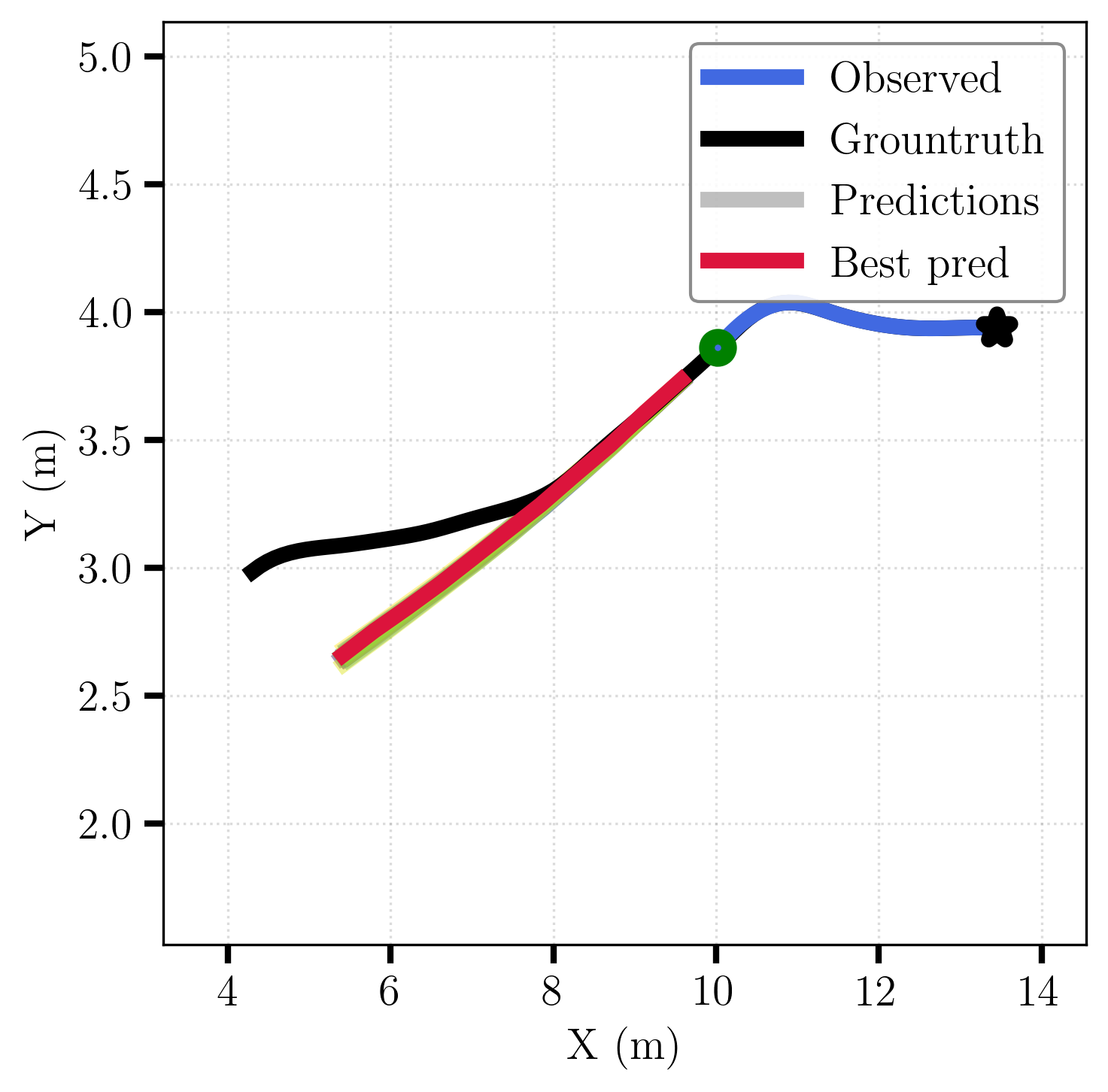} &
        \includegraphics[width=0.19\textwidth]{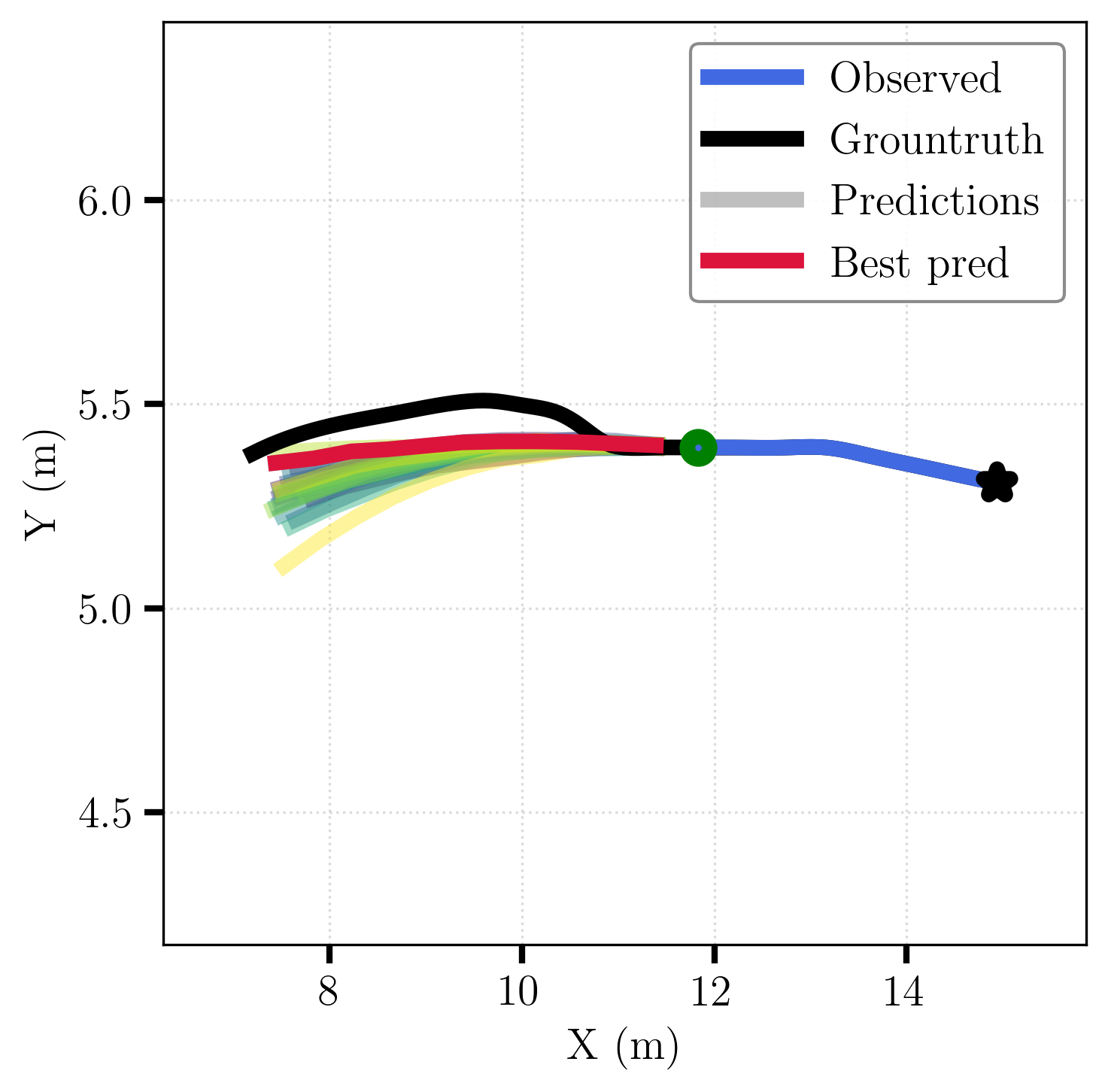} &
        \includegraphics[width=0.19\textwidth]{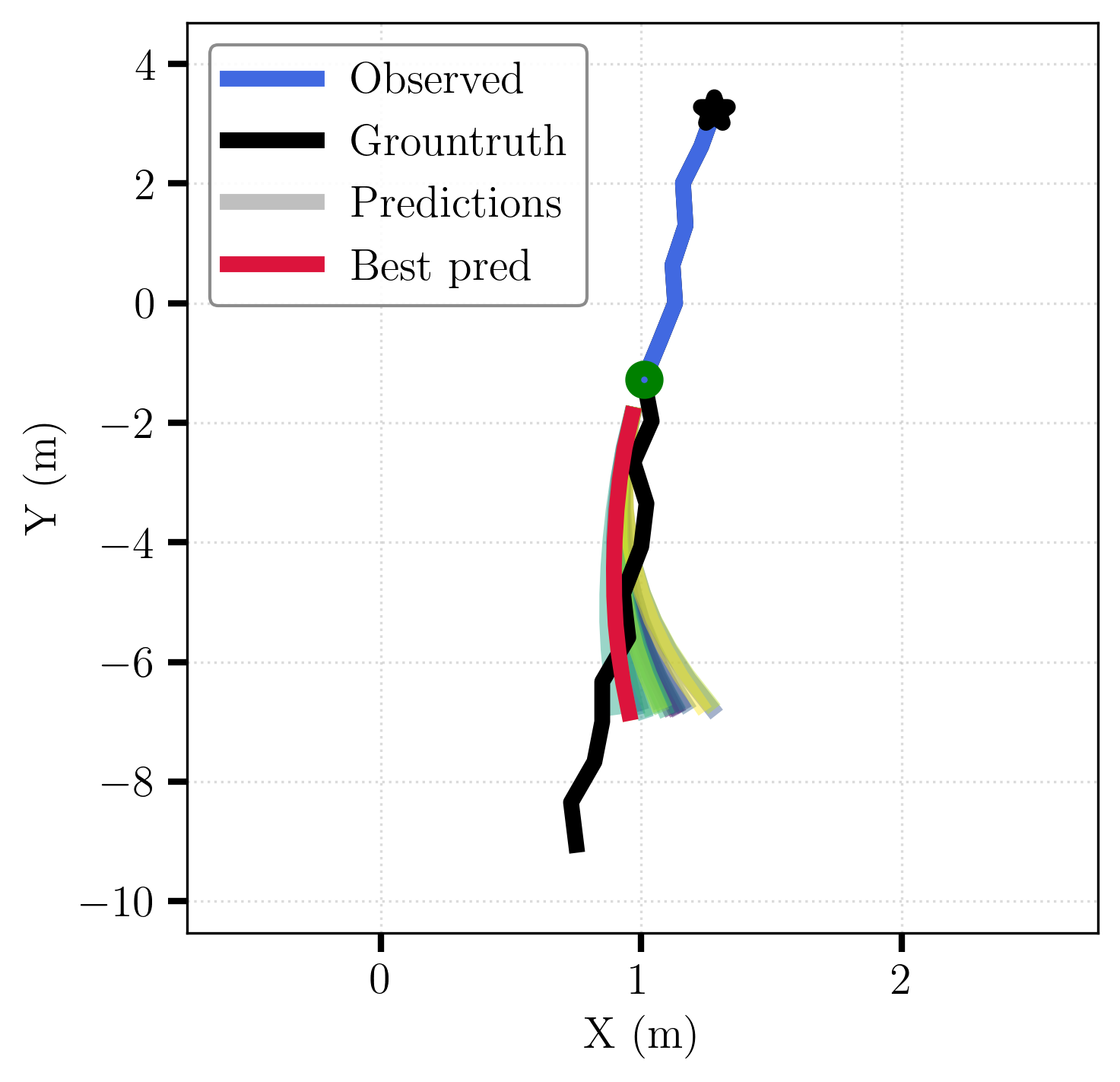} \\
        \includegraphics[width=0.19\textwidth] {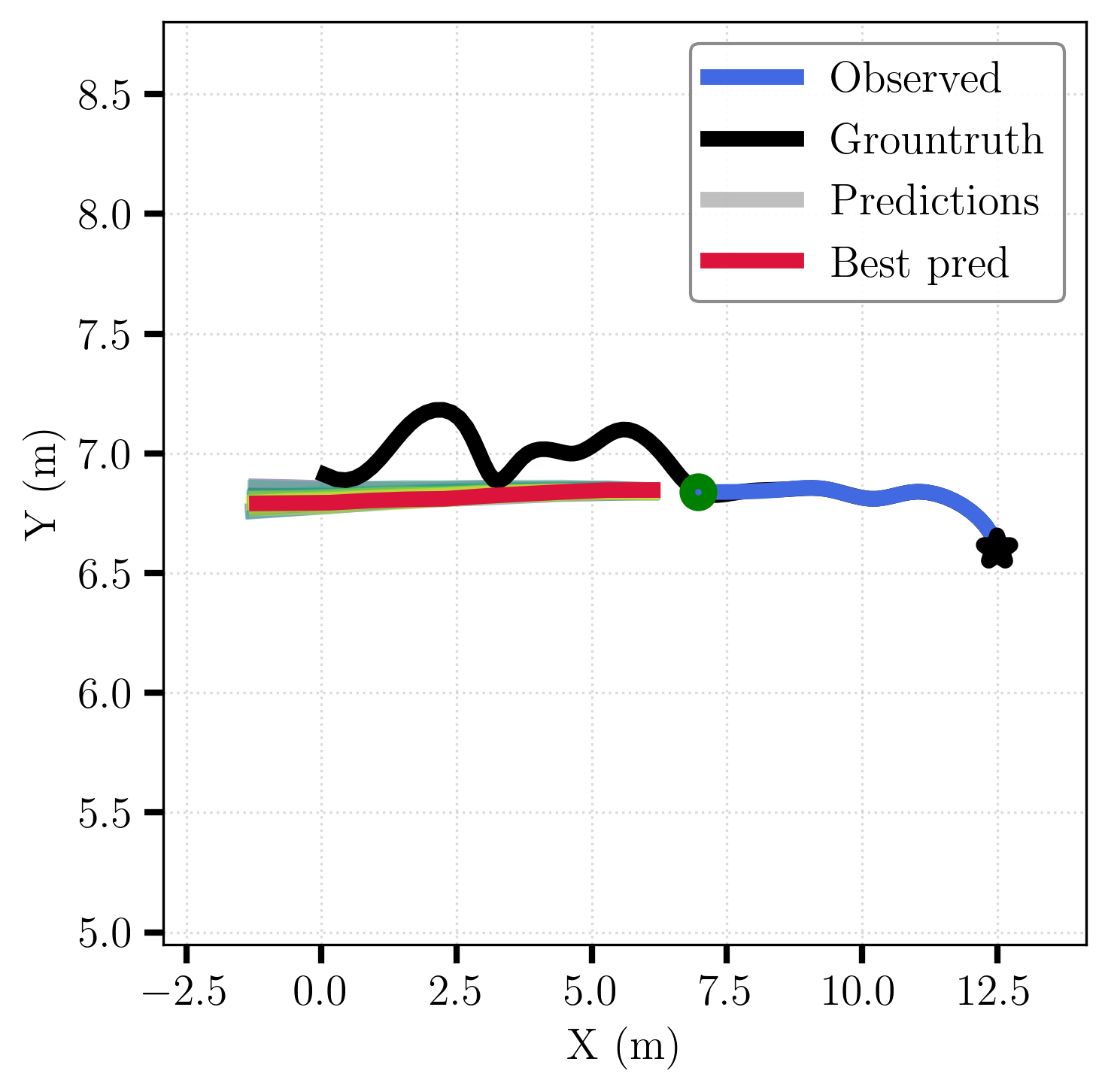} &
        \includegraphics[width=0.19\textwidth]{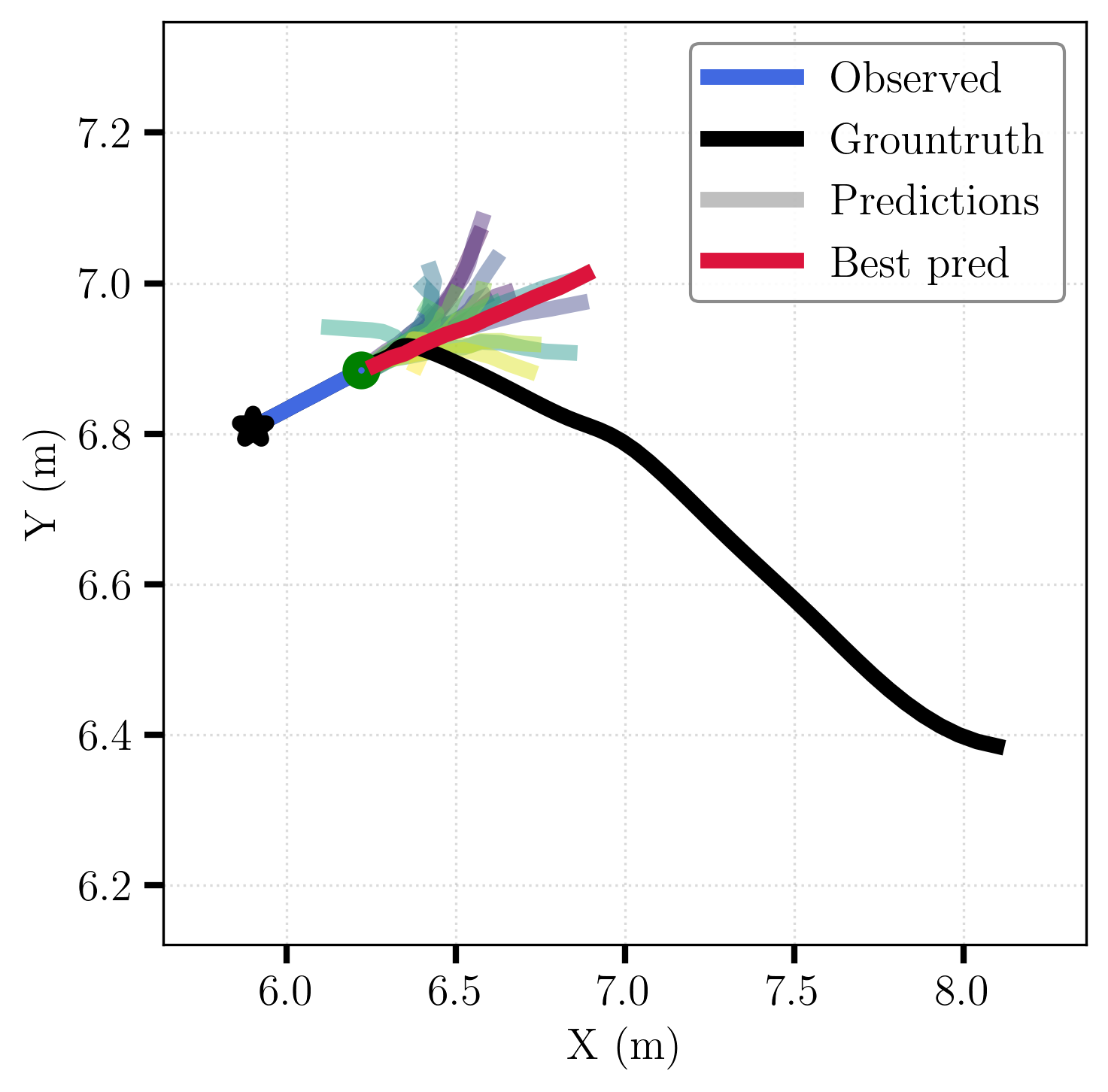} &
        \includegraphics[width=0.19\textwidth]{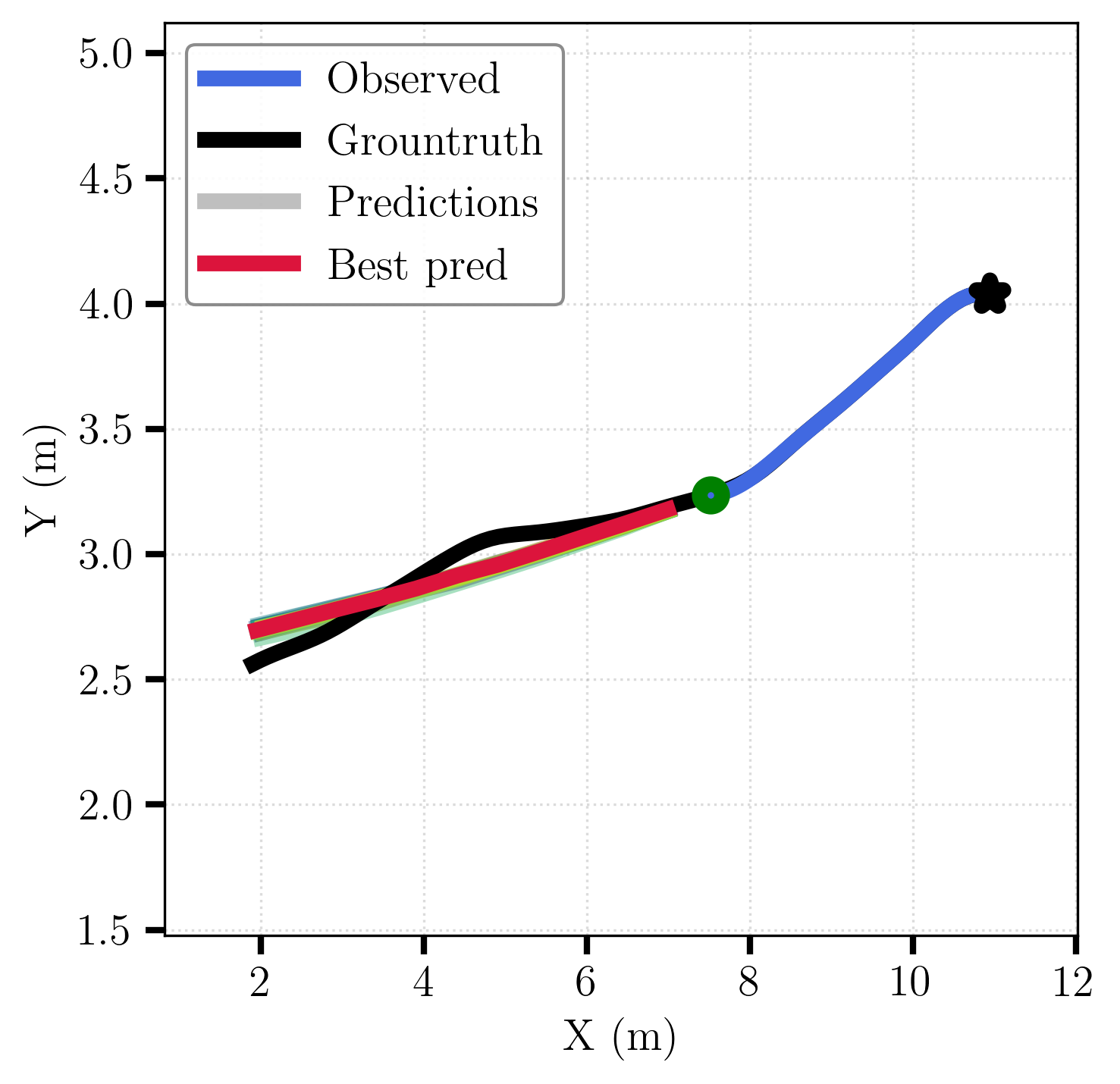} &
        \includegraphics[width=0.19\textwidth]{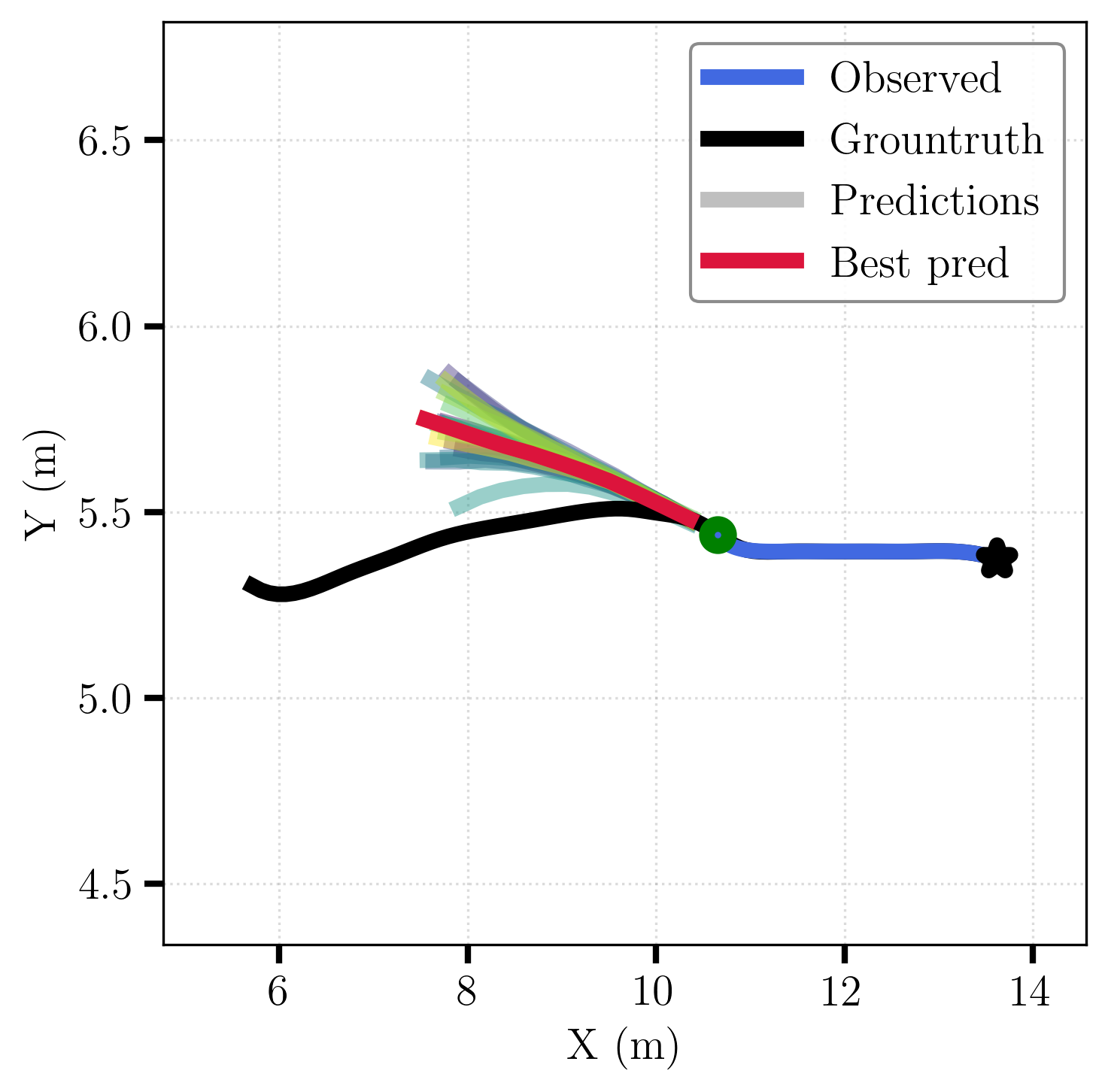} &
        \includegraphics[width=0.19\textwidth]{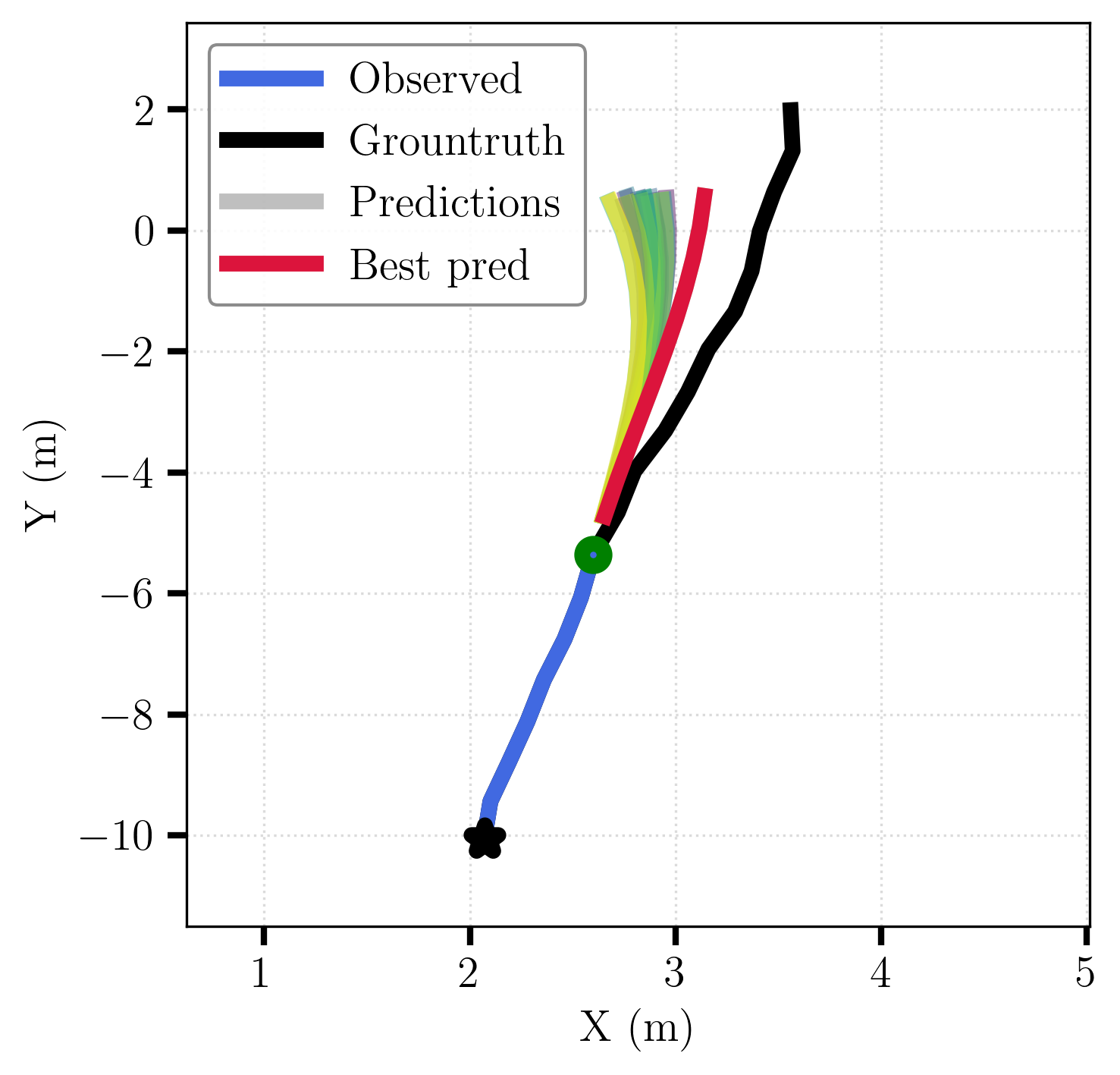} \\
        \includegraphics[width=0.19\textwidth] {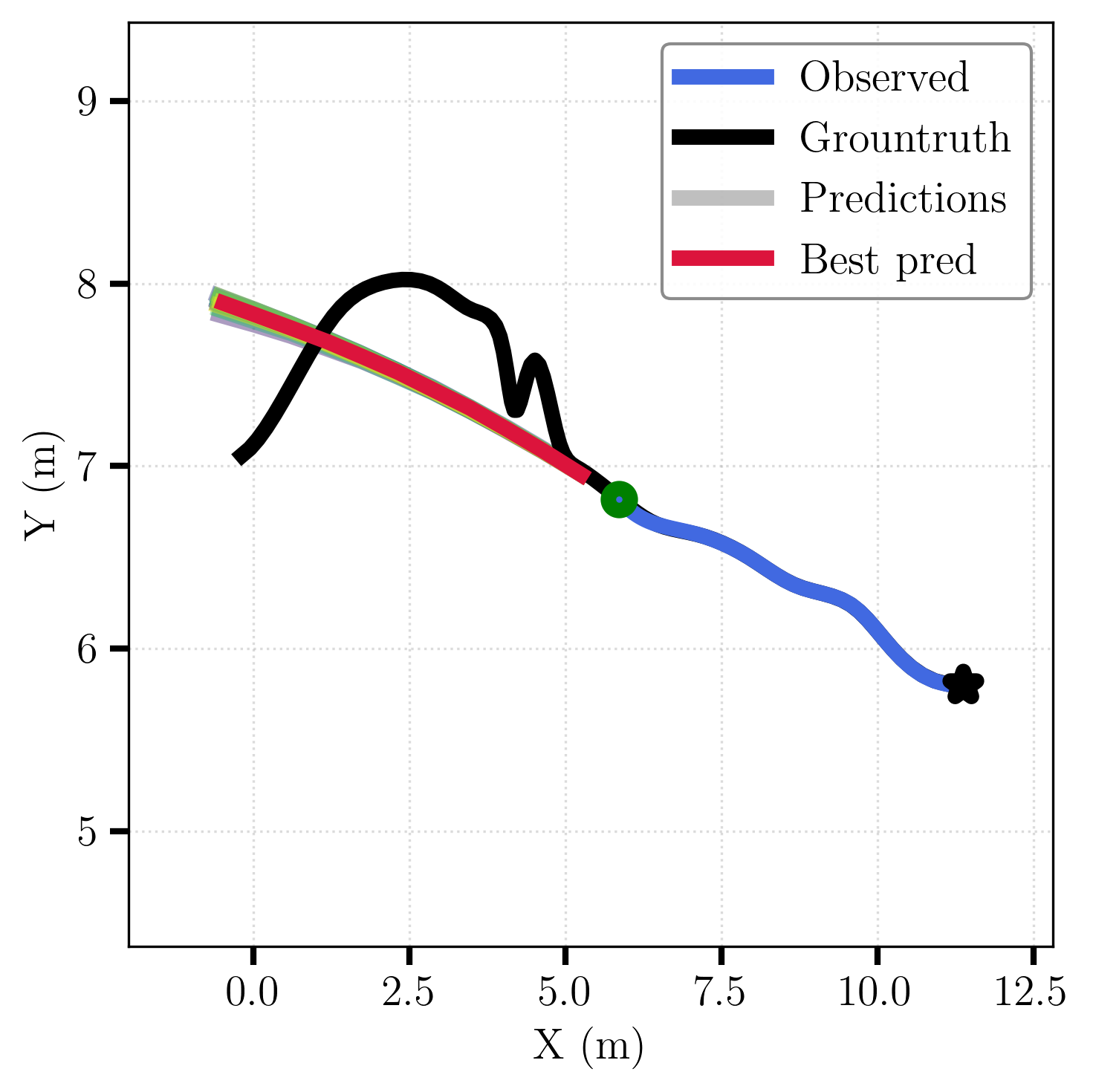} &
        \includegraphics[width=0.19\textwidth]{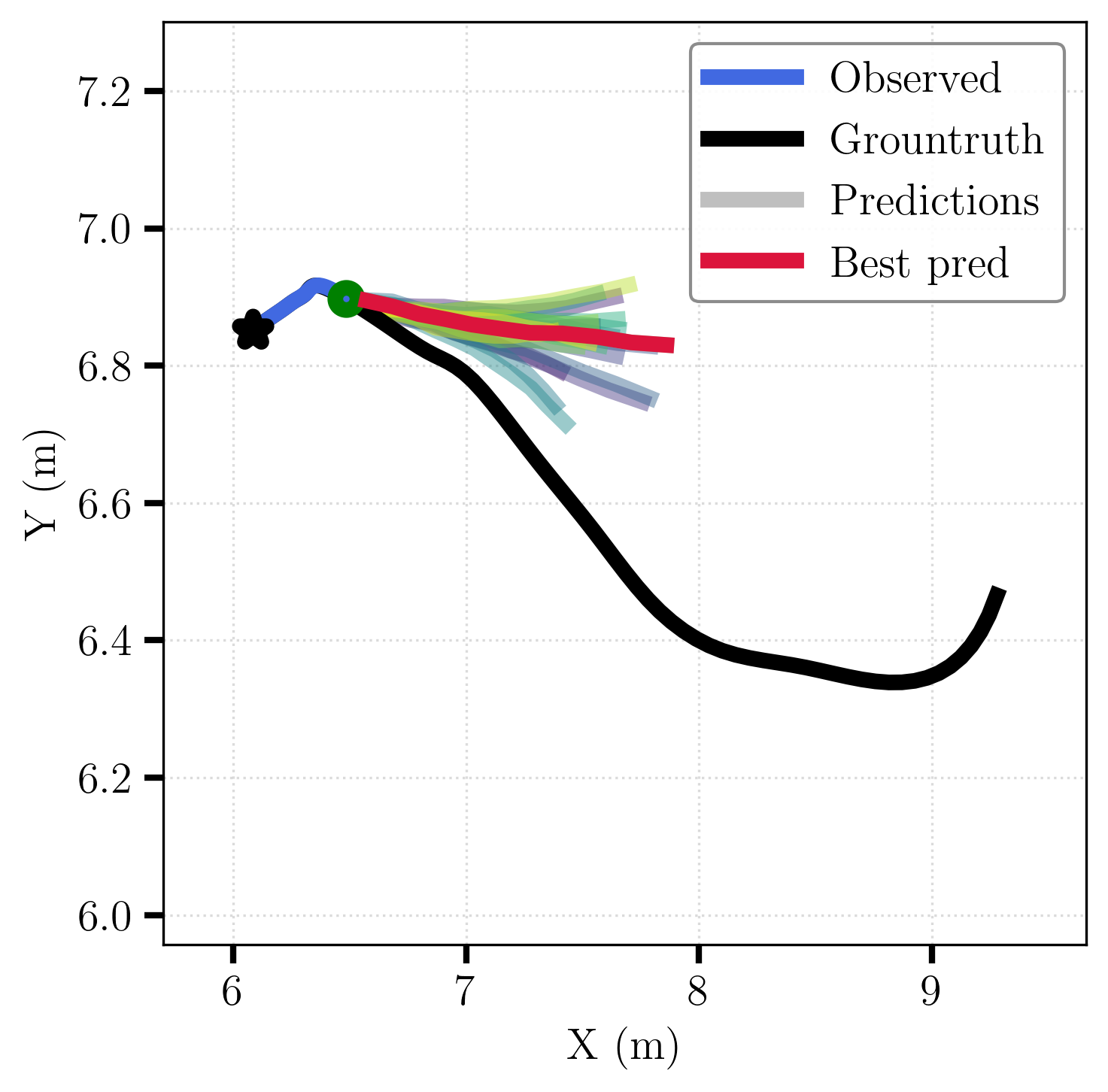} &
        \includegraphics[width=0.19\textwidth]{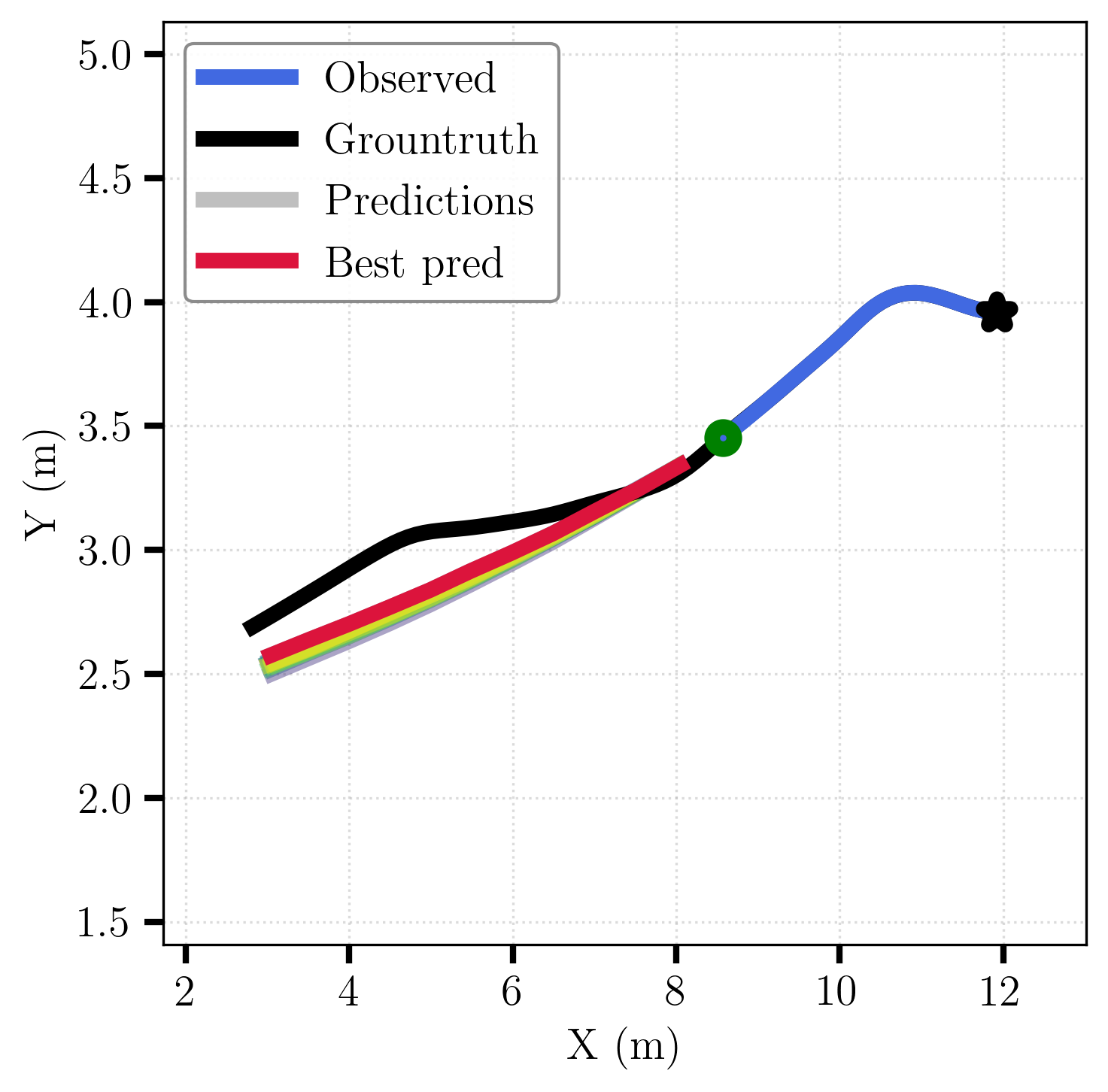} &
        \includegraphics[width=0.19\textwidth]{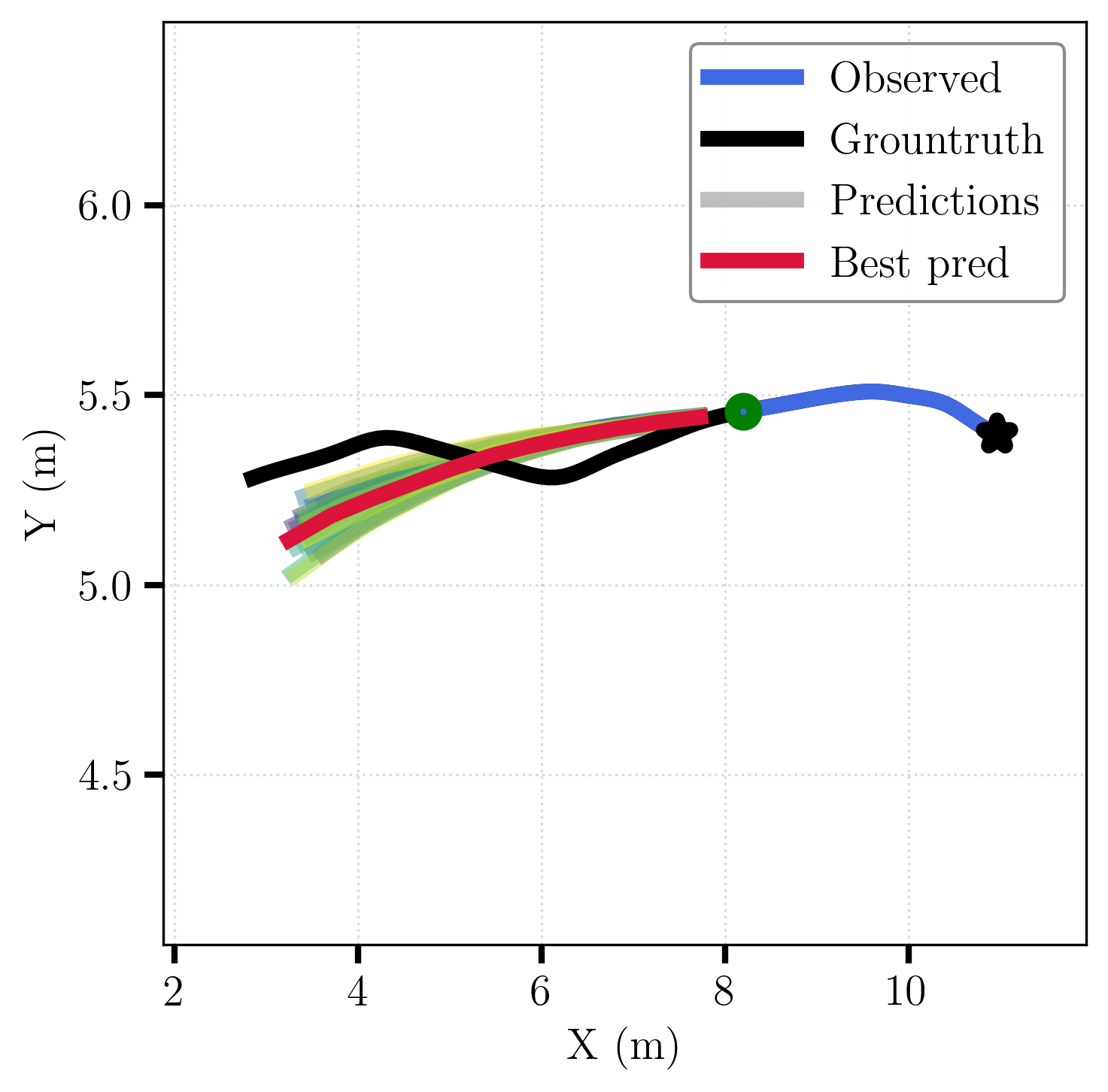} &
        \includegraphics[width=0.19\textwidth]{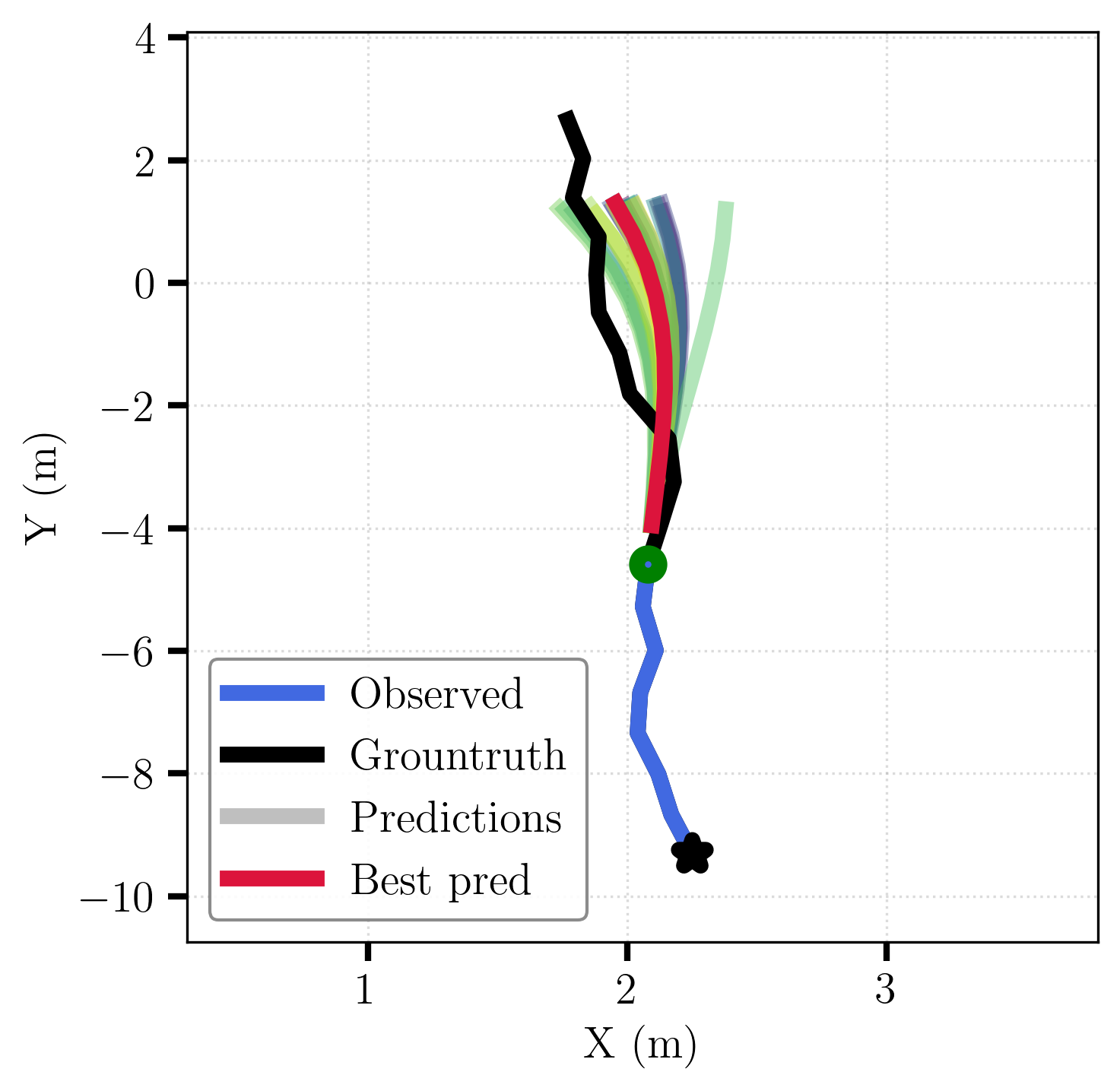} \\
        \includegraphics[width=0.18\textwidth] {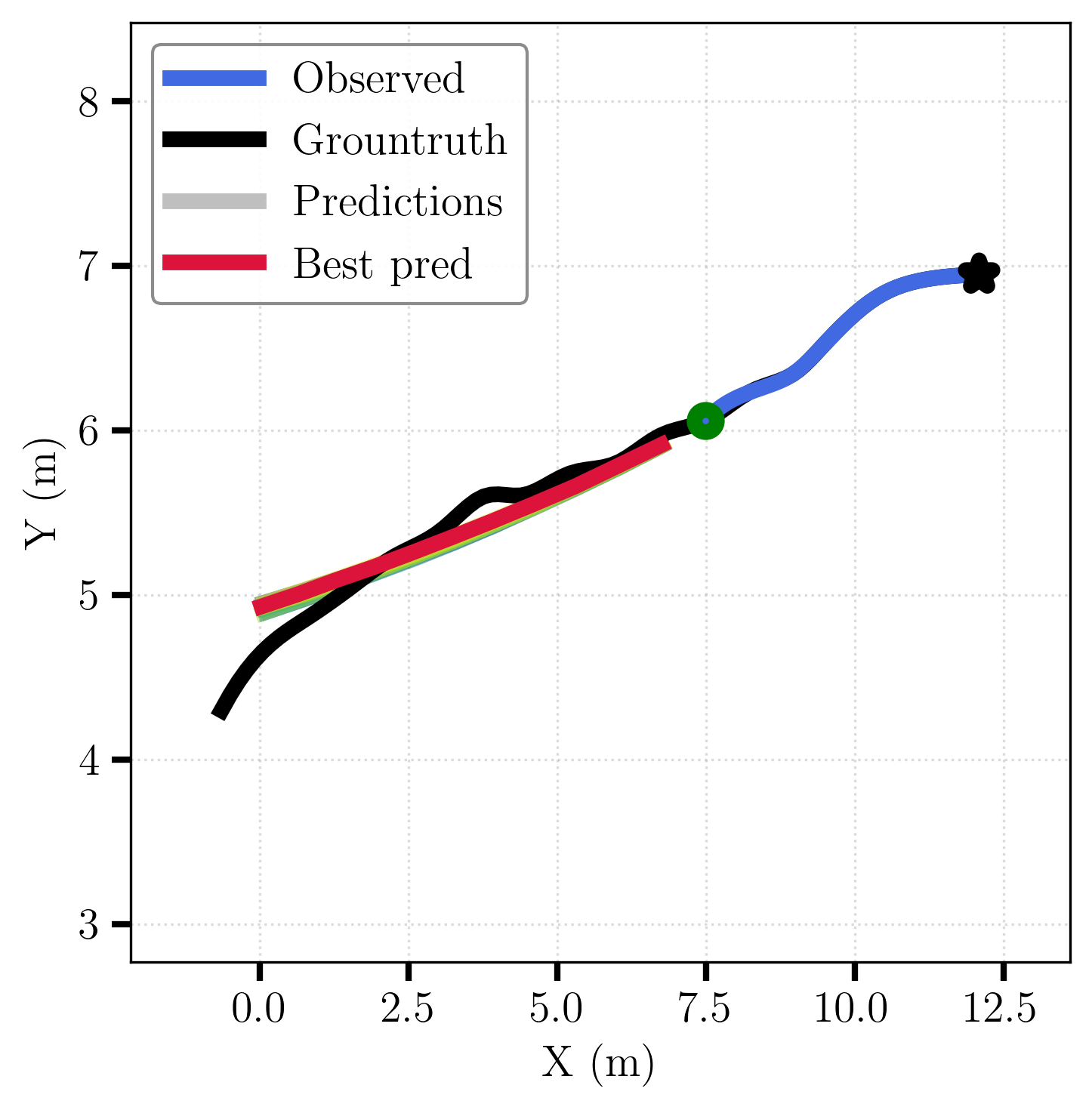} &
        \includegraphics[width=0.19\textwidth]{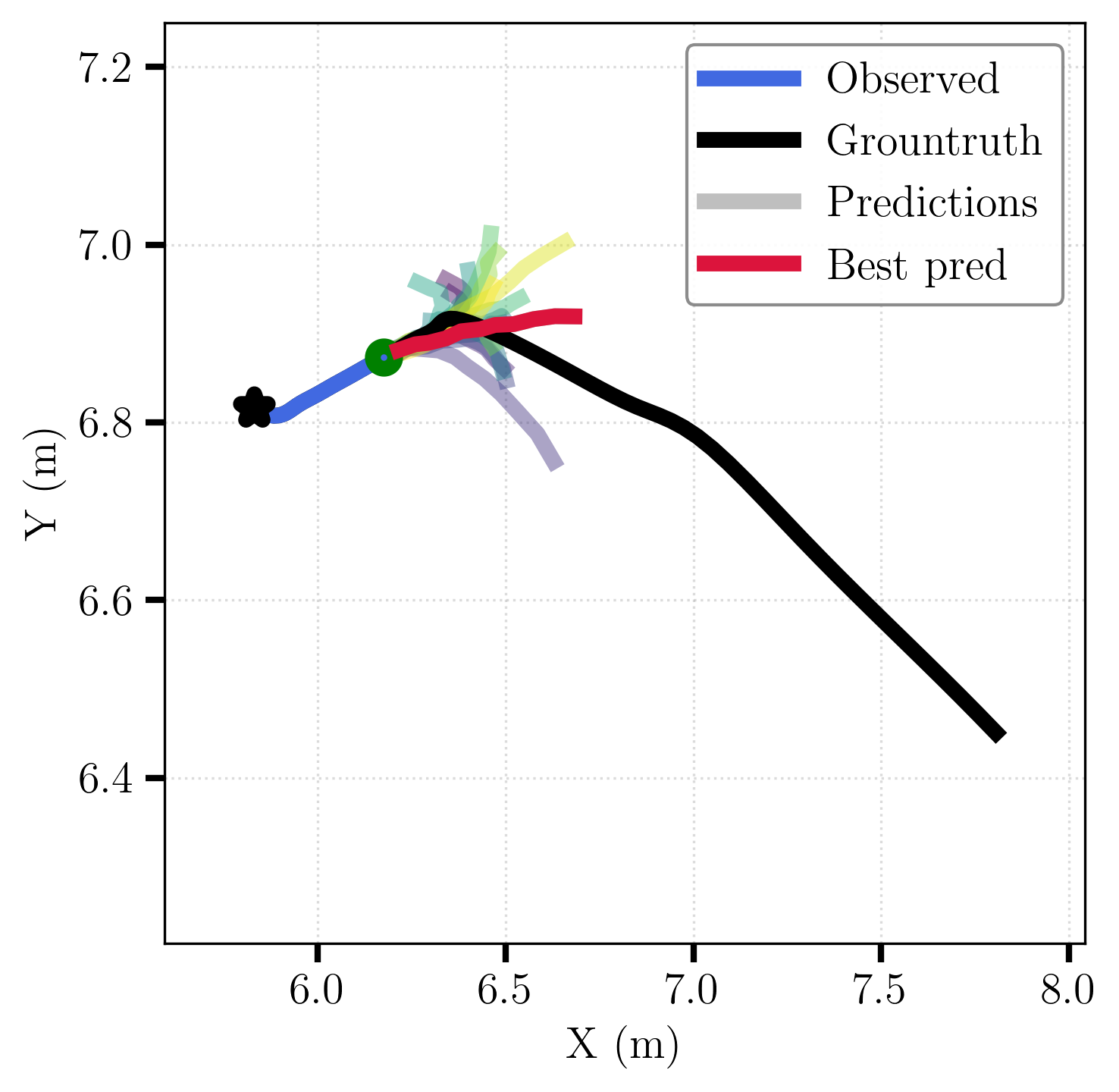} &
        \includegraphics[width=0.19\textwidth]{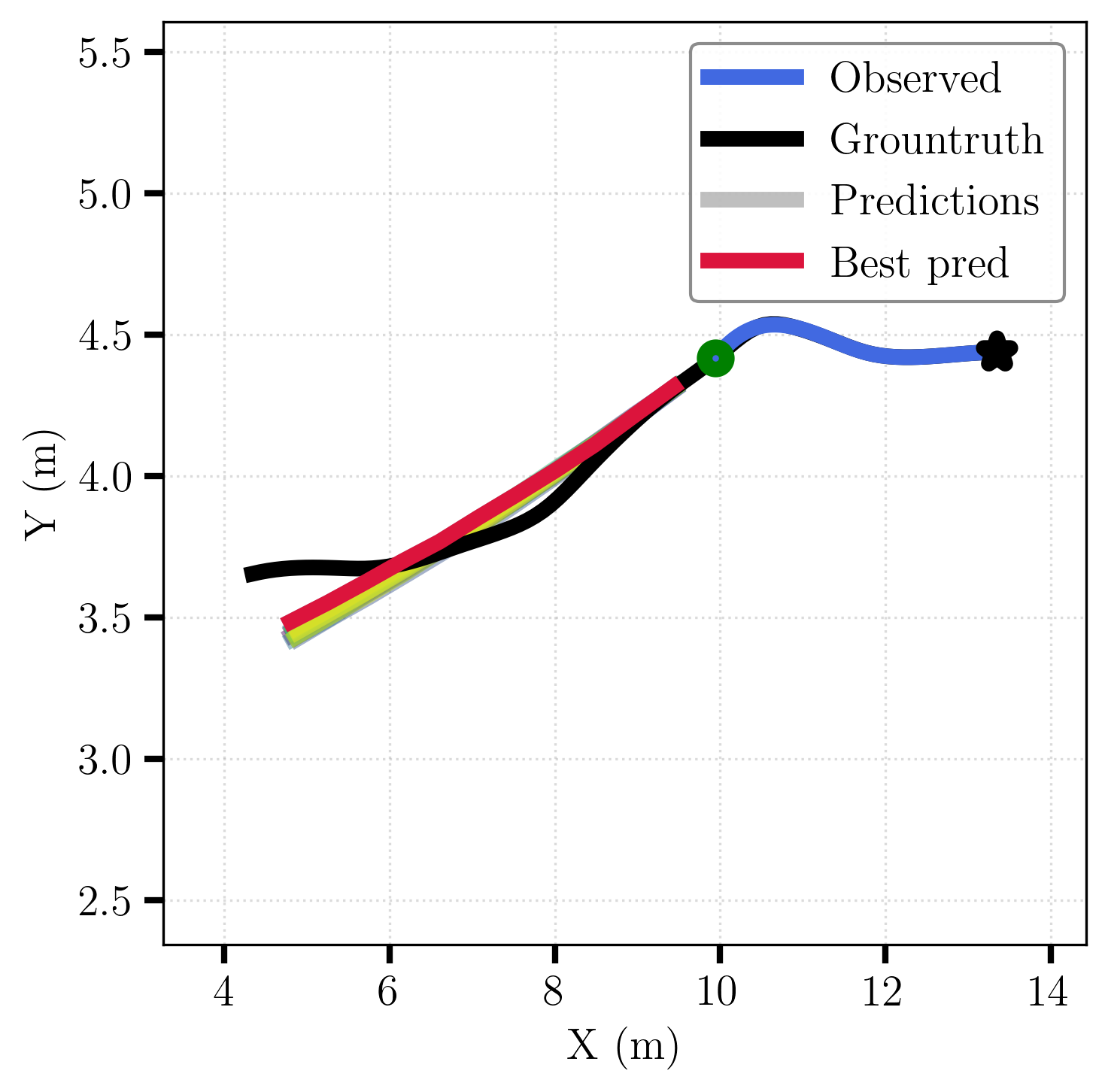} &
        \includegraphics[width=0.19\textwidth]{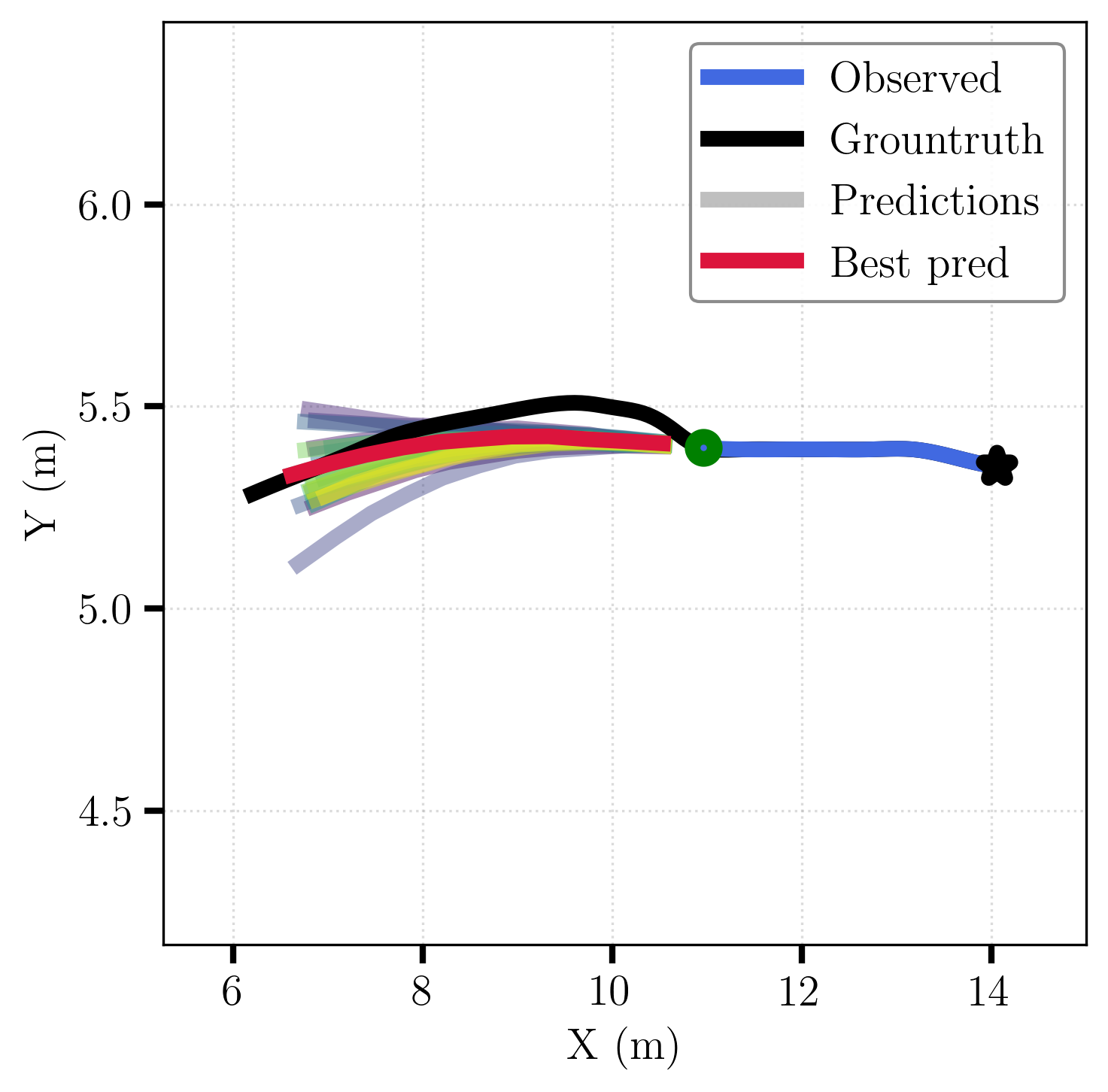} &
        \includegraphics[width=0.19\textwidth]{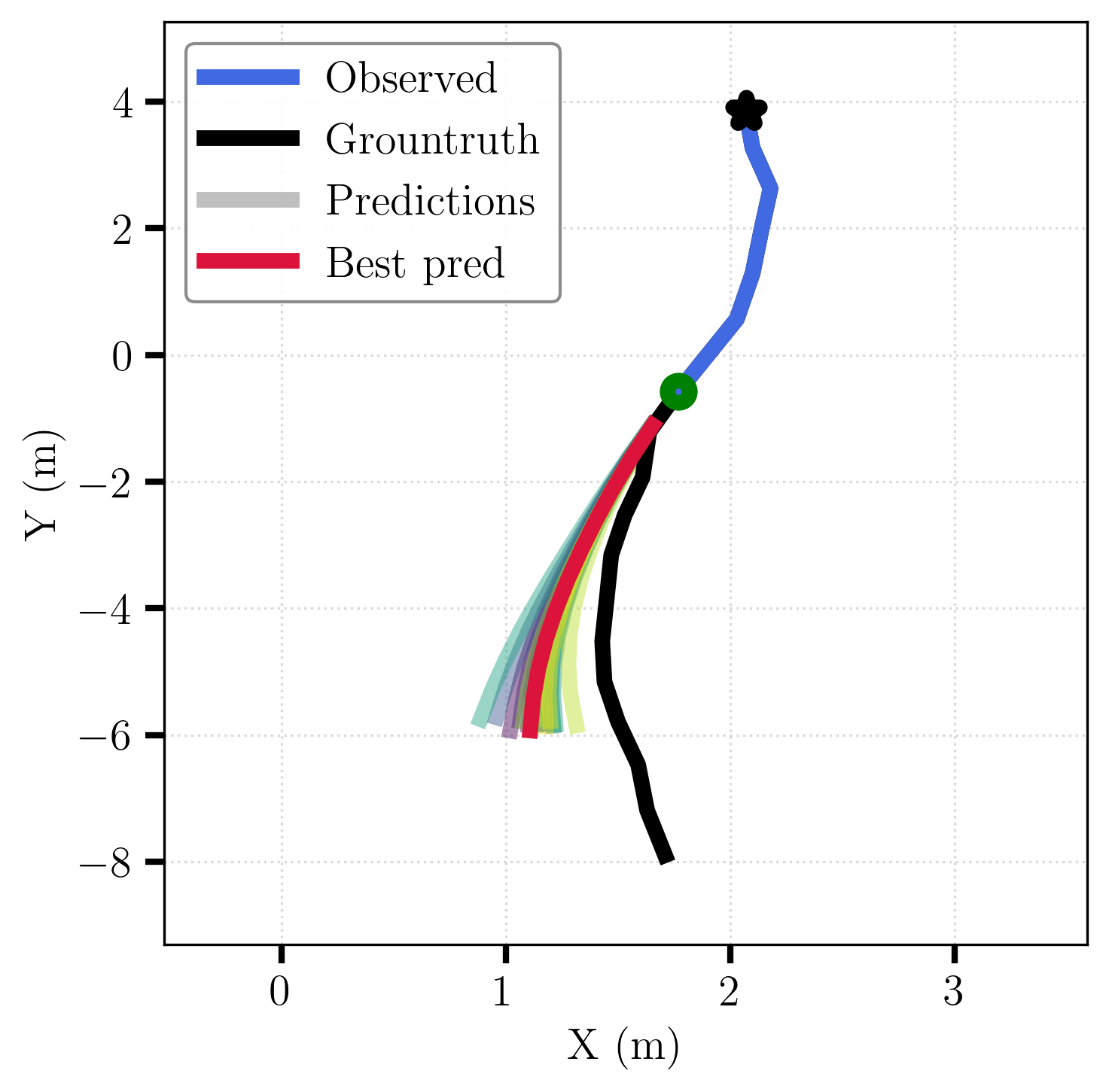} \\
    \end{tabular}
    \caption{Prediction Visualization on the ETH/UCY benchmark:
    Each column corresponds to one scene (ETH, UNIV, Zara1, Zara2, Hotel) and
    each row shows a different sample. Blue: observed history; green star: start;
    black: ground-truth future; red: best prediction; thin colored
    curves: the spread of sampled predictions.}
    \label{fig:evaluation}
\end{figure*}
\subsection{Training Details}
\label{sec:training}
We optimize with AdamW (learning rate $3\times10^{-4}$, weight decay
$10^{-4}$, $\beta=(0.9,0.98)$) under a cosine-annealing-with-warm-restarts
schedule ($T_0{=}30$, $T_{\text{mult}}{=}2$, $\eta_{\min}{=}10^{-6}$) for up to
$150$ epochs. The objective is a composite loss combining a Huber term
($\delta{=}0.1$) with an $L_1$ penalty (weight $0.1$) on normalized velocities:
\begin{equation}
\mathcal{L} = \text{Huber}_{\delta}(\hat{\mathbf{v}},\mathbf{v})
            + \lambda\,\lVert \hat{\mathbf{v}} - \mathbf{v}\rVert_1,
\quad \lambda = 0.1 .
\label{eq:loss}
\end{equation}
Gradients are clipped to a max norm of $1.0$. Light data augmentation
(velocity scaling $\pm10\%$ and axis flips) is applied on the training split
only. The data is split $85/15$ for train/validation with a fixed seed
($42$).

\begin{figure*}[htbp]
    \centering

    \includegraphics[width=0.19\textwidth]{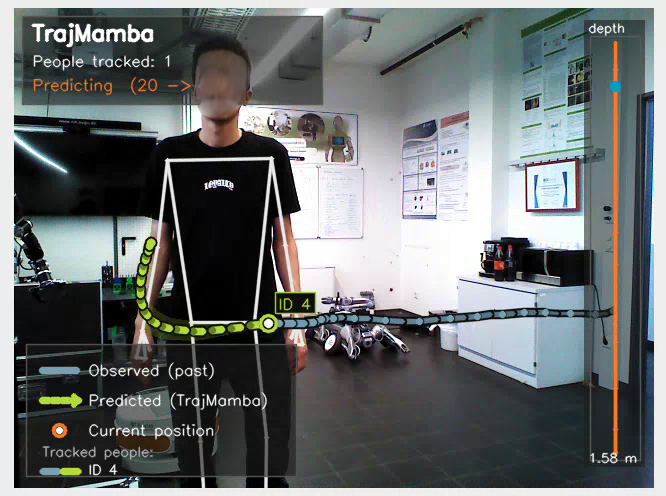}\hfill
    \includegraphics[width=0.19\textwidth]{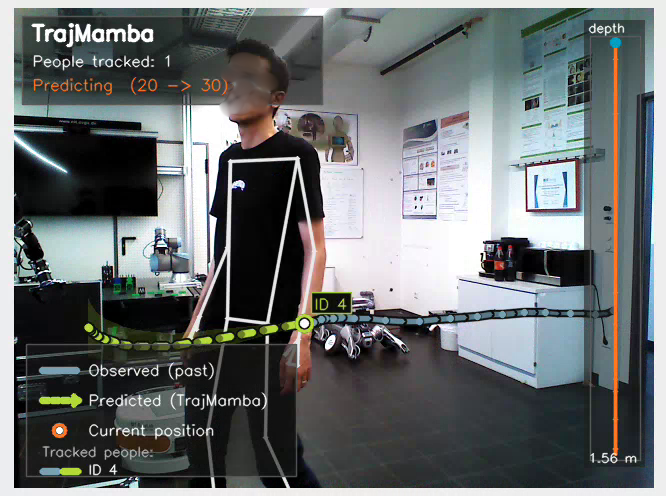}\hfill
    \includegraphics[width=0.19\textwidth]{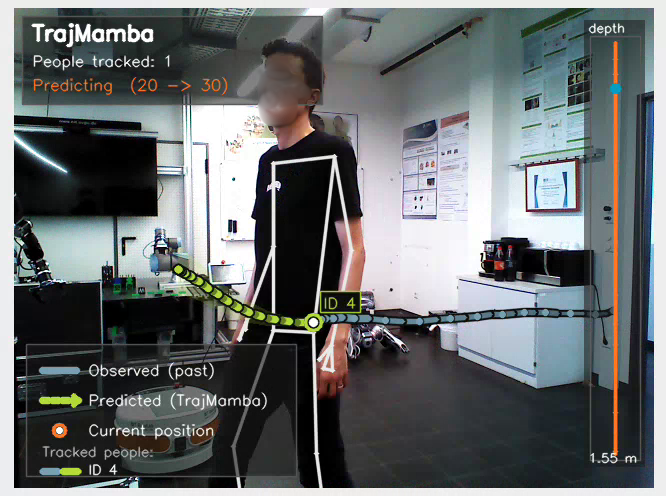}\hfill
    \includegraphics[width=0.19\textwidth]{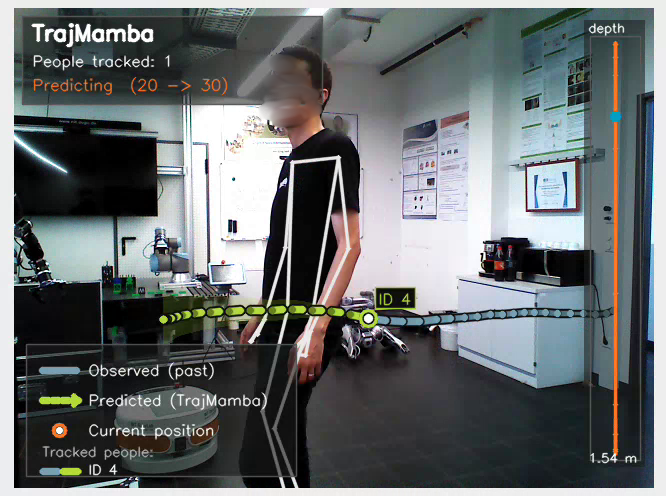}\hfill
    \includegraphics[width=0.19\textwidth]{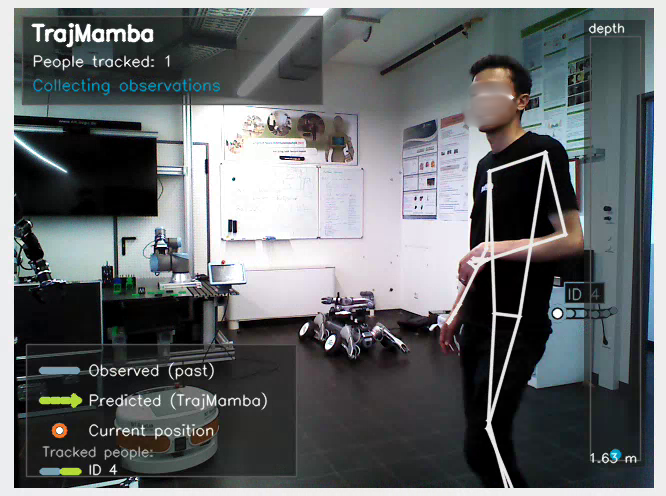}

    \vspace{0.2em}
    \begin{tikzpicture}
        \draw[-{Latex[length=3mm]}, line width=1pt, gray]
            (0.85\textwidth,0) -- (0.13\textwidth,0)
            node[midway, above=2pt, black] {\small Right to Left};
    \end{tikzpicture}
    \vspace{0.6em}

    \includegraphics[width=0.19\textwidth]{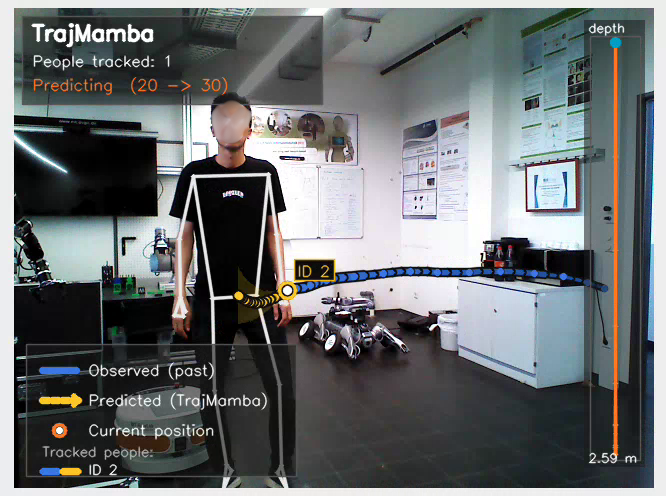}\hfill
    \includegraphics[width=0.19\textwidth]{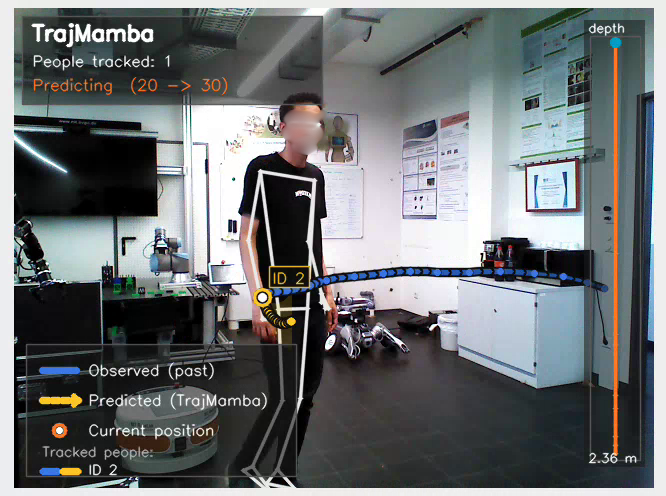}\hfill
    \includegraphics[width=0.19\textwidth]{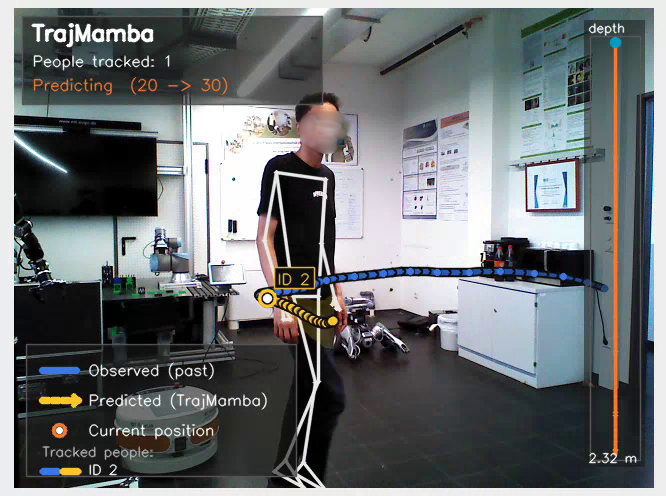}\hfill
    \includegraphics[width=0.19\textwidth]{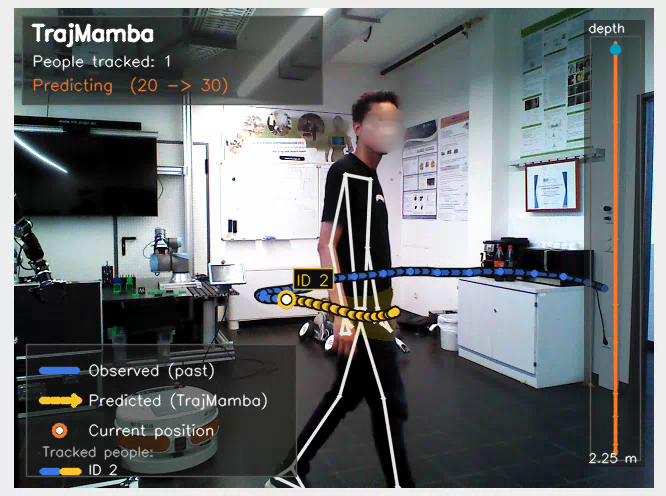}\hfill
    \includegraphics[width=0.19\textwidth]{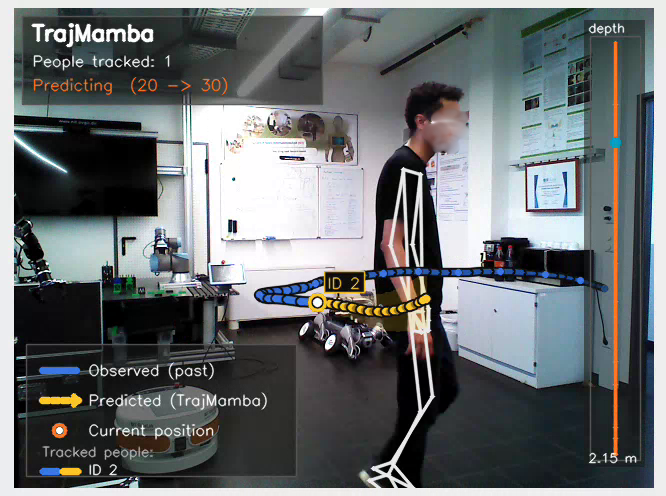}

    \vspace{0.2em}
    \begin{tikzpicture}
        \draw[-{Latex[length=3mm]}, line width=1pt, gray]
            (0.13\textwidth,0) -- (0.85\textwidth,0)
            node[midway, above=2pt, black] {\small Left to Right};
    \end{tikzpicture}
    \vspace{0.6em}

    \includegraphics[width=0.19\textwidth]{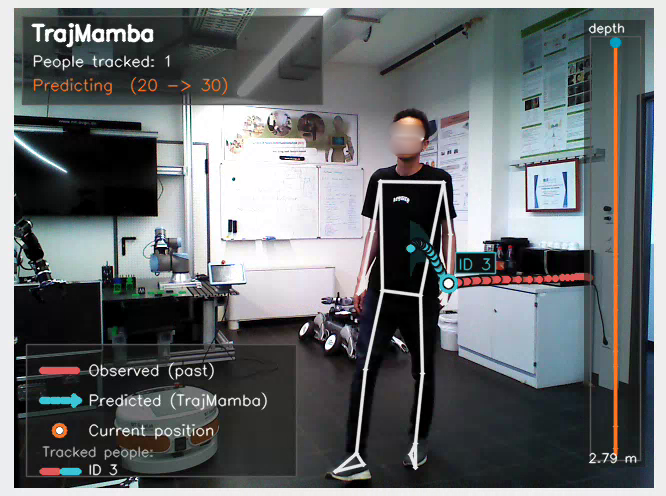}\hfill
    \includegraphics[width=0.19\textwidth]{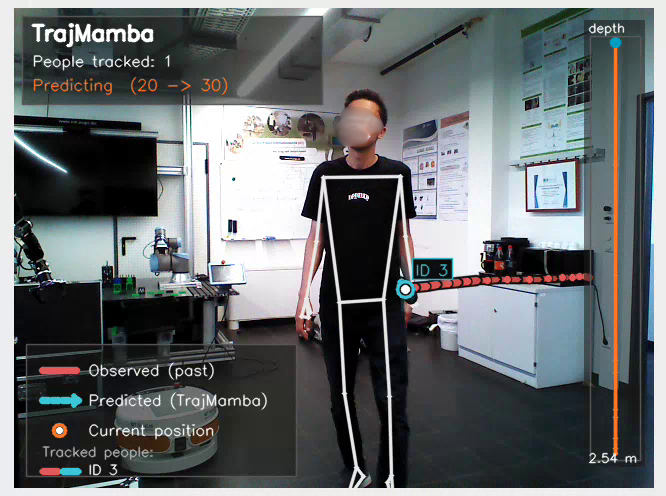}\hfill
    \includegraphics[width=0.19\textwidth]{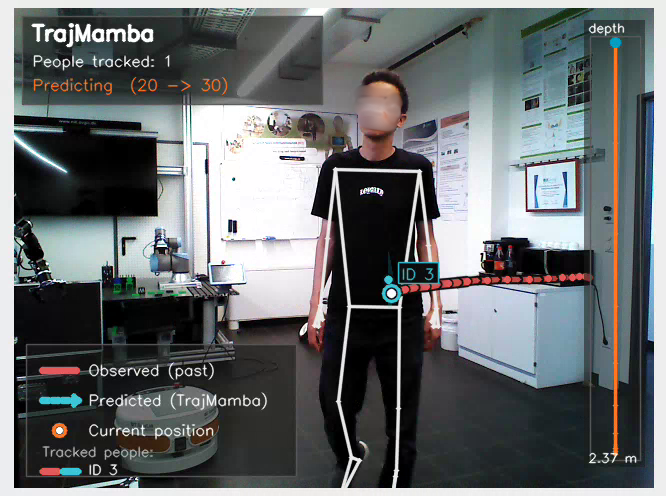}\hfill
    \includegraphics[width=0.19\textwidth]{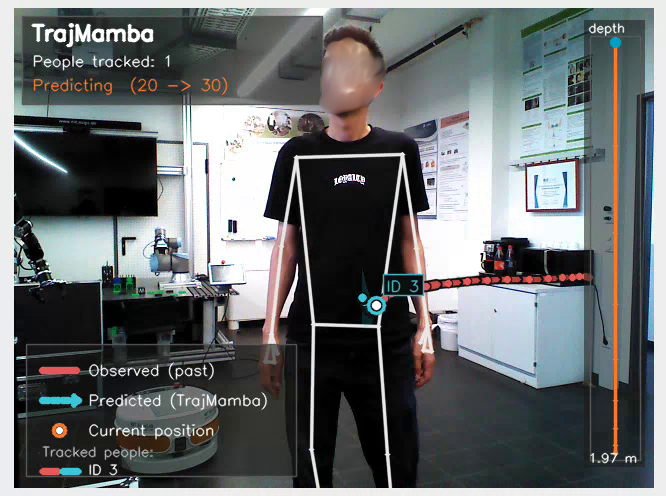}\hfill
    \includegraphics[width=0.19\textwidth]{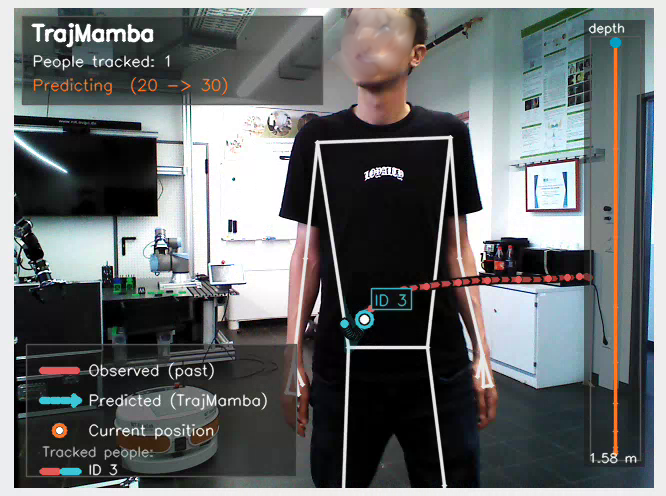}

    \vspace{0.2em}
    \begin{tikzpicture}
        \draw[-{Latex[length=3mm]}, line width=1pt, gray]
            (0.49\textwidth,0.25) -- (0.49\textwidth,-0.25)
            node[midway, right=2pt, black] {\small approaching};
    \end{tikzpicture}

    \caption{Real Time prediction across three motion patterns.
    Top: Person motion right$\rightarrow$left of the Robot. Middle: Person motion
    left$\rightarrow$right of the Robot. Bottom: Person motion toward the camera. }
    \label{fig:selected_frames}
\end{figure*}
\section{Experiments and Practical Evaluation}
\label{sec:experiments}

\subsection{Datasets}
RGB-D Bonn Person Tracking: This dataset contains indoor
RGB-D sequences recorded with an Asus Xtion Pro Live camera. It features a single person performing dynamic motions (straight walks, sudden stops, direction changes) with the sensor held near chest height, close to the geometry of a mobile robot camera viewpoint. The raw sequence used in this work consists of 580 RGB and 580 depth frames. After processing with a MediaPipe PoseLandmarker for 3D keypoint extraction, followed by Kalman and EMA smoothing and downsampling to 10\,Hz, the resulting trajectories are segmented into overlapping temporal windows. It produces 145 trajectory windows in total, which are split into 124 training and 21 validation windows using an 85/15 split.

We additionally evaluate the model on the standard
ETH/UCY benchmark under the leave-one-scene-out protocol with
$T_\text{obs}{=}8$(observed frames), $T_\text{pred}{=}12$(predicted frames). The positions are read directly from the benchmark annotation files, so the RGB-D pose/depth/filtering pipeline does not
apply. The remainder of the pipeline is unchanged: from the observed positions, we form the
current velocities by finite differences $\mathbf{v}_t = \mathbf{p}_{t+1} -
\mathbf{p}_t$ and standardize them; the network predicts the future velocities
$\hat{\mathbf{v}}_t$; and the future trajectory is recovered by integrating
these velocities from the last observed position,
$\hat{\mathbf{p}}_{t+1} = \hat{\mathbf{p}}_t + \hat{\mathbf{v}}_t$.

\subsection{Evaluation Metrics}
We use the two standard metrics, the average displacement error (ADE) and the final displacement error (FDE):
\begin{align}
\text{ADE} &= \frac{1}{N T_p} \sum_{i=1}^{N} \sum_{t=1}^{T_p}
        \left\|\hat{\mathbf{p}}_t^{(i)} - \mathbf{p}_t^{(i)}\right\|_2,
\label{eq:ade}\\
\text{FDE} &= \frac{1}{N}     \sum_{i=1}^{N}
        \left\|\hat{\mathbf{p}}_{T_p}^{(i)}
              - \mathbf{p}_{T_p}^{(i)}\right\|_2,
\label{eq:fde}
\end{align}
where $N$ represents the total number of motion sequences in the test set, and $T_p = T_\text{pred}$ represents the prediction horizon, while $\mathbf{p}$ and $\hat{\mathbf{p}}$ denote the true and predicted positions, respectively. The errors are computed in meters on the integrated positions.

\begin{table*}[t]
\centering
\caption{Comparison with state of the art methods on ETH/UCY benchmark (ADE\,/\,FDE in
meters).}
\label{tab:eth_ucy_comparison}
\renewcommand{\arraystretch}{1.15}
\setlength{\tabcolsep}{5pt}
\small
\begin{tabular}{lccccccccc}
\toprule
\textbf{Dataset} &
\makecell[c]{\textbf{BR-GAN} \\ \cite{pang2022br}} &
\makecell[c]{\textbf{Social-GAN} \\ \cite{gupta2018social}} &
\makecell[c]{\textbf{CNN} \\ \cite{nikhil2018convolutional}} &
\makecell[c]{\textbf{Trajectron} \\ \cite{ivanovic2019trajectron}} &
\makecell[c]{\textbf{Social-STGCNN} \\ \cite{mohamed2020social}} &
\makecell[c]{\textbf{STGAT}\\ \cite{huang2019stgat}} &
\makecell[c]{\textbf{SR-LSTM} \\ \cite{zhang2019sr}} &
\makecell[c]{\textbf{DTGAN} \\ \cite{xie2024pedestrian}} &
\makecell[c]{\textbf{(Ours)}} \\
\midrule
ETH      &0.73 / 1.37 & 0.81 / 1.52 & 1.04 / 2.07 & 0.59 / 1.14 & 0.64 / 1.11 & 0.57 / 0.89 & 0.63 / 1.25 & 0.68 / 1.43 & 0.55 / 0.85 \\
UNIV     & 0.53 / 1.07 & 0.60 / 1.26 & 0.57 / 1.21 & 0.54 / 1.13 & 0.44 / 0.79 & 0.49 / 0.96 & 0.51 / 1.10 & 0.51 / 1.07 & 0.31 / 0.58 \\
ZARA1    & 0.35 / 0.71 & 0.34 / 0.69 & 0.43 / 0.90 & 0.43 / 0.83 & 0.34 / 0.53 & 0.29 / 0.59 & 0.41 / 0.90 & 0.31 / 0.67 & 0.22 / 0.38 \\
ZARA2    & 0.35 / 0.72 & 0.42 / 0.84 & 0.34 / 0.75 & 0.43 / 0.85 & 0.30 / 0.48 & 0.24 / 0.54 & 0.32 / 0.70 & 0.28 / 0.59 & 0.17 / 0.31 \\
HOTEL    & 0.55 / 1.13 & 0.72 / 1.61 & 0.59 / 1.27 & 0.35 / 0.66 & 0.49 / 0.85 & 0.41 / 0.86 & 0.37 / 0.74 & 0.30 / 0.52 & 0.20 / 0.43 \\
\midrule
\textbf{AVG} & 0.50 / 1.00 & 0.58 / 1.18 & 0.59 / 1.24 & 0.47 / 0.92 & 0.44 / 0.75 & 0.40 / 0.77 & 0.45 / 0.94 & 0.42 / 0.85 & \textbf{0.29 / 0.51} \\
\bottomrule
\end{tabular}
\end{table*}
\subsection{Visual Analysis of Trajectory Prediction}        
Figure~\ref{fig:evaluation} shows qualitative predictions across the five
ETH/UCY scenes, with one sample per column. In each panel the blue curve is the
observed history, the green star marks its start, the black dashed curve is the ground truth, and the red curve is the model's prediction, all aligned. On the mostly
straight motions in Zara1, Zara2, and Hotel, the prediction closely follows the ground truth. In ETH and UNIV, where the person turns within the horizon, the model gets the initial direction right but rounds off the turn,
predicting a gentler curve than the sharp change that actually happens. This is expected from a model that commits to a single trajectory rather than hedging across several possible futures.
\subsection{Comparison with State of the Art Methods}

Table~\ref{tab:eth_ucy_comparison} compares our model against eight established
pedestrian-prediction methods on the ETH/UCY benchmark under the standard
leave one scene out protocol, with deterministic ($K{=}1$) predictions. Across
all five scenes, our model attains the lowest errors, reaching ADE of 0.29\,m and FDE of 0.51\,m, among strongest baselines such as STGAT ($0.40/0.77$) and Social-STGCNN ($0.44/0.75$).  The improvement is consistent rather than scene-specific. This supports the view that capturing both causal and anti-causal temporal context aids short-horizon motion understanding without the cost load, while keeping the model compact enough for embedded platforms.
\subsection{Real-Time On-Robot Implementation}
\label{sec:realtime}

To verify our model, we deployed the
complete system on a PAL Robotics TIAGo mobile robot,
using its head-mounted RGB-D camera for the color and depth streams and an
NVIDIA Jetson Orin is the on-board computer that runs the entire pipeline. The
ROS2 node subscribes directly to the image topics, and reuses the exact same
pipeline-consistent module described in Section~\ref{sec:extraction}: each
synchronized RGB-D pair is passed through MediaPipe pose extraction, $7\times7$
depth back-projection, and the Kalman/EMA filtering. After which the
self-contained saved checkpoint predicts the future velocities and integrates them
into a $3$\,s forward trajectory. 
 
Figure~\ref{fig:selected_frames} shows the live overlay across the three motion patterns relative to the robot: lateral motion from right to
left (top), left to right (middle), and approaching motion toward the camera
(bottom). In every case, the observed history (solid trail) and the predicted
future path (dashed arrow) is rendered directly on the camera stream together
with the current position and a depth gauge, allowing the predicted heading
and the approaching/receding behavior along the optical axis to be read at a glance.

\section{Conclusion}
\label{sec:conclusion}
We presented a lightweight bidirectional state space model for short human trajectory prediction. It encodes a window of observed velocities with stacked bidirectional Mamba blocks and decodes future velocities with the GRU head and recovers positions by velocity integration. On the standard ETH/UCY benchmark, it reached the 
lowest average error among eight baselines, with an average ADE of
$0.29$\,m and FDE of $0.51$\,m. The model is small enough for real-time deployment, which we confirm by running on a TIAGo robot with an on-board Jetson Orin. Currently it tracks several people but predicts each independently, so modeling the social interactions between them is the next step. Building on this, we plan to integrate TrajMamba into a social-aware navigation system, where its predictions let the robot anticipate human motion and move safely through shared, crowded spaces.

\bibliographystyle{IEEEtran}
\bibliography{ref}

\end{document}